\documentclass[10pt,twocolumn,letterpaper]{article}

\usepackage[pagenumbers]{cvpr} % To force page numbers, e.g. for an arXiv version

\usepackage{titletoc}
\usepackage{tocloft}
\usepackage{algorithm}
\usepackage{algpseudocode}

\definecolor{cvprblue}{rgb}{0.21,0.49,0.74}
\usepackage[pagebackref,breaklinks,colorlinks,allcolors=cvprblue]{hyperref}

\def\paperID{19459} % *** Enter the Paper ID here
\def\confName{CVPR}
\def\confYear{2026}

\title{MoECLIP: Patch-Specialized Experts for Zero-shot Anomaly Detection}

\author{Jun Yeong Park\qquad
JunYoung Seo\qquad 
Minji Kang \qquad 
Yu Rang Park\thanks{Corresponding author.}
\\
Yonsei University, Seoul, Korea
\\
{\tt\small pjk000011@yonsei.ac.kr seojy@yonsei.ac.kr mnzzy@yonsei.ac.kr yurangpark@yuhs.ac}
}
\usepackage{booktabs} 
\usepackage{multirow} 
\usepackage{graphicx} 
\usepackage{amsmath}  
\usepackage{colortbl}
\usepackage{multirow}
\usepackage{wrapfig}
\usepackage{tabularx}
\usepackage{makecell}
\usepackage{enumitem}
\usepackage{marvosym}
\usepackage[accsupp]{axessibility} 
\usepackage{float} 
\usepackage{pifont}
\usepackage{multirow}
\usepackage{booktabs}
\usepackage{array}
\usepackage{adjustbox}

\definecolor{lightgreen}{RGB}{240,255,255}
\definecolor{darkred}{RGB}{180,0,0}
\definecolor{lightblue1}{RGB}{100, 149, 237} 
\definecolor{lightgreen1}{RGB}{34, 139, 34} 
\definecolor{lightorange1}{RGB}{255, 165, 0} 

\begin{document}
\maketitle
\begin{abstract}
The CLIP model's outstanding generalization has driven recent success in Zero-Shot Anomaly Detection (ZSAD) for detecting anomalies in unseen categories. The core challenge in ZSAD is to specialize the model for anomaly detection tasks while preserving CLIP's powerful generalization capability. Existing approaches attempting to solve this challenge share the fundamental limitation of a patch-agnostic design that processes all patches monolithically without regard for their unique characteristics. To address this limitation, we propose \textbf{MoECLIP}, a Mixture-of-Experts (MoE) architecture for the ZSAD task, which achieves patch-level adaptation by dynamically routing each image patch to a specialized Low-Rank Adaptation (LoRA) expert based on its unique characteristics. Furthermore, to prevent functional redundancy among the LoRA experts, we introduce (1) Frozen Orthogonal Feature Separation (FOFS), which orthogonally separates the input feature space to force experts to focus on distinct information, and (2) a simplex equiangular tight frame (ETF) loss to regulate the expert outputs to form maximally equiangular representations. Comprehensive experimental results across 14 benchmark datasets spanning industrial and medical domains demonstrate that MoECLIP outperforms existing state-of-the-art methods. The code is available at \url{https://github.com/CoCoRessa/MoECLIP}.
\end{abstract}    
\vspace{-2.5mm}
\section{Introduction}
\label{sec:intro}
\begin{figure}[tb]
	\centering
	\includegraphics[width=0.92\columnwidth]{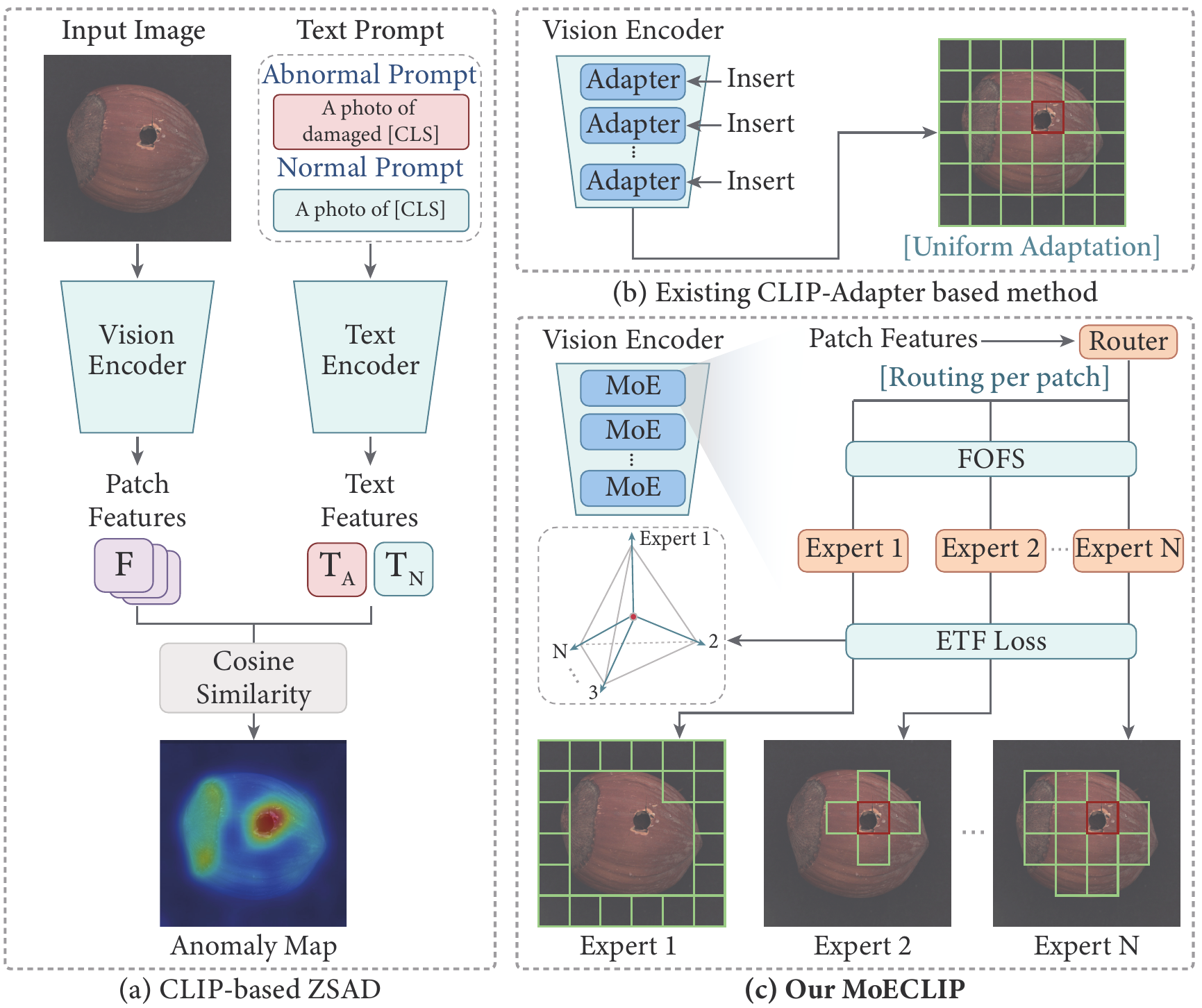}
    \vspace{-0.7em}
	\caption{\textbf{Comparison between existing CLIP-Adapter based method and our MoECLIP.} (a) The general CLIP-based ZSAD framework. (b) Existing methods apply a uniform adaptation to all patches, regardless of their unique characteristics. (c) In contrast, our MoECLIP utilizes a Mixture of Experts to achieve patch-specialized adaptation, dynamically routing each patch to experts that are differentiated by FOFS and an ETF loss.}
	\label{Fig1}
\vspace{-1.0em}
\end{figure}
Visual Anomaly Detection (AD)~\cite{cao2024survey} aims to identify anomalous regions that deviate from normal patterns and serves as a critical technology in various fields, including industrial defect detection~\cite{mvctec, deng2022anomaly, adformer}, medical image diagnosis~\cite{ccalli2021deep,fernando2021deep}. Due to the scarcity of anomalous samples, collecting labeled anomaly data is impractical, making Unsupervised Anomaly Detection (UAD) the traditional AD paradigm~\cite{defard2021padim,wyatt2022anoddpm,roth2022towards,guo2025dinomaly,he2024learning,he2024mambaad}, where models learn only from normal data to detect anomalies. However, UAD is still predicated on the availability of sufficient normal data, posing a significant constraint in data-scarce environments.

As an alternative to overcome these limitations, the Zero-Shot Anomaly Detection (ZSAD) paradigm~\cite{jeong2023winclip,chen2023april,zhou2023anomalyclip,cao2024adaclip,ma2025aa,qu2025bayesian} has emerged, leveraging the rich visual-semantic understanding of Vision-Language Models (VLMs) like Contrastive Language-Image Pretraining (CLIP)~\cite{radford2021learning} to enhance generalization for unseen classes. As illustrated in \cref{Fig1}(a), the core idea of CLIP-based ZSAD is to detect anomalies by measuring the similarity between patch embeddings and text embeddings. However, CLIP is pretrained for global semantic understanding, making it suboptimal for detecting localized anomalies~\cite{li2024promptad}. Thus, a key challenge is adapting CLIP for anomaly detection while preserving its powerful generalization capability.

Motivated by this challenge, recent ZSAD methods have attempted to enhance image patch representations for the anomaly detection task. PromptAD~\cite{li2024promptad} and AnomalyCLIP~\cite{zhou2023anomalyclip} seek to focus on local regions by replacing CLIP's QKV Attention with V-V Attention. Parameter-Efficient Fine-Tuning (PEFT)~\cite{han2024parameter} methods, such as AdaCLIP~\cite{cao2024adaclip} and AA-CLIP~\cite{ma2025aa}, introduce prompting Tokens or Adapters to improve anomaly detection performance while preserving CLIP's generalization capability. However, these methods share a limitation: as depicted in \cref{Fig1}(b), they follow a patch-agnostic design that performs uniform adaptation across all patches, without considering that different image regions represent distinct structures or semantics (e.g., object components, backgrounds).

To address these limitations, we propose \textbf{MoECLIP}, a novel framework that utilizes Patch-Specialized Experts. The core idea of MoECLIP, as depicted in \cref{Fig1}(c), is to strategically integrate a Mixture of Experts (MoE)~\cite{shazeer2017outrageously} module into the CLIP Vision Encoder, which dynamically routes each image patch to the most suitable expert based on its unique characteristics. Specifically, to preserve CLIP's generalization capability and avoid overfitting on auxiliary datasets, MoECLIP adopts a PEFT-based feature adaptation approach, where MoE experts are implemented as lightweight Low-Rank Adaptation (LoRA)~\cite{hu2022lora} modules. However, a naive ensemble of LoRA experts can lead to functional redundancy, where experts learn similar functions. To prevent this and promote expert specialization, we employ two complementary strategies: Frozen Orthogonal Feature Separation (FOFS) enforces expert separation at the input stage by orthogonally separating the feature space, ensuring non-overlapping feature subspaces by construction, while at the output stage, a simplex equiangular tight frame (ETF)~\cite{papyan2020prevalence} loss regularizes expert outputs to follow an equiangular target structure, ensuring clear differentiation. Through this design, MoECLIP overcomes the limitations of existing uniform adaptation methods, enabling fine-grained patch-level adaptation and enhancing generalization performance across diverse datasets.

Our key contributions are summarized as follows:
\begin{enumerate}
    \item \textit{\textbf{Pioneering a MoE-based architecture for Zero-Shot Anomaly Detection.}} We are the first to introduce an approach to the ZSAD task that dynamically routes each image patch to a specialized expert, establishing a new paradigm of patch-level adaptation for this task.
    \item \textit{\textbf{Novel Mechanisms for Expert Specialization.}} To prevent functional redundancy and boost differentiation among LoRA experts, we introduce Frozen Orthogonal Feature Separation (FOFS), which orthogonally separates the input feature space, and a simplex equiangular tight frame (ETF) loss that enforces an equiangular structure on expert outputs.
    \item \textit{\textbf{State-of-the-art performance on comprehensive benchmark datasets.}} In comprehensive experiments on 14 benchmark datasets spanning industrial and medical domains, MoECLIP achieves state-of-the-art (SOTA) performance in both the anomaly classification and segmentation tasks of ZSAD.
\end{enumerate}

\section{Related Work}
\subsection{Zero-Shot Anomaly Detection (ZSAD)}
ZSAD, where generalization performance on unseen classes is critical, has made significant strides based on the CLIP model. WinCLIP~\cite{jeong2023winclip}, the first study to apply CLIP to ZSAD, proposes a method for calculating similarity between handcrafted text prompts and multi-scale image patches. Subsequent research has attempted various approaches to better adapt CLIP for the anomaly detection task. April-GAN~\cite{chen2023april} and CLIP-AD~\cite{chen2024clip} seek to enhance CLIP's patch representations by introducing a linear adapter. AnomalyCLIP~\cite{zhou2023anomalyclip} and FiLo~\cite{gu2024filo} utilize a prompt learning approach, introducing learnable parameters to the text prompt. Bayes-PFL~\cite{qu2025bayesian} treats the text prompt space as a learnable probability distribution from a Bayesian inference perspective. AdaCLIP~\cite{cao2024adaclip} and VCP-CLIP~\cite{qu2024vcp} use a hybrid prompt approach to enhance the alignment of text and patch embeddings. AA-CLIP~\cite{ma2025aa} combines a loss function to increase the separability of normal and abnormal text embeddings with a Residual Adapter. In this way, there have been various attempts in the ZSAD field to improve text and image patch representations. However, all approaches to enhancing image patch representations share a common limitation: they apply the same, monolithic transformation to all patches, without considering their unique individual characteristics. This patch-agnostic approach fundamentally undermines the model's ability to identify fine-grained anomaly patterns.

\subsection{Mixture of Experts (MoE)}
Mixture-of-Experts (MoE)~\cite{shazeer2017outrageously,li2024mixlora} is a conditional computation architecture wherein a gating network dynamically activates only a small subset of experts best suited for a given input, enabling massive-scale expansion and excellent generalization performance. Owing to these properties, MoE has been a key strategy for building large-scale models~\cite{lepikhin2020gshard,du2022glam,fedus2022switch,jiang2024mixtral,guo2025deepseek}, and its application has recently expanded to the field of reconstruction-based Unsupervised Anomaly Detection (UAD)~\cite{meng2024moead,gu2025anomalymoe}. However, two significant gaps remain. First, the application of MoE to the generalization-critical ZSAD task remains entirely unexplored. Second, existing methods to MoE's functional redundancy problem~\cite{liu2023diversifying,cui2025cmoa} typically attempt to enforce differentiation by applying output-level constraints, such as contrastive loss~\cite{feng2025comoe} or orthogonality regularization~\cite{feng2025omoe}. However, these methods are limited as they fail to address feature overlap at both the input and output stages of experts. Our MoECLIP is the first to address both gaps: we introduce MoE to the ZSAD task to solve the patch-agnostic problem, and ensure robust expert specialization by applying FOFS and the ETF loss to control differentiation at both the input and output stages of the LoRA experts.
\section{Preliminaries}
\subsection{Low-Rank Adaptation (LoRA)}
\textit{Low-Rank Adaptation (LoRA)}~\cite{hu2022lora} is a Parameter-Efficient Fine-Tuning (PEFT) technique that freezes the pretrained weight matrix 
${W}_0 \in \mathbb{R}^{d_{1} \times d_{2}}$ and injects a trainable, low-rank update $\Delta W = BA$, where down projection matrix $A \in \mathbb{R}^{r \times d_2}$ and up-projection matrix $B \in \mathbb{R}^{d_1 \times r}$ are the trainable matrices. The final adapted weight is computed as $W = W_0 + \Delta W$. Since the rank $r$ satisfies $r \ll \min(d_1, d_2)$, this method significantly reduces the number of trainable parameters compared to simple linear adapters and reduces the risk of overfitting.

\subsection{Mixture of Experts (MoE)}
A \textit{Mixture-of-Experts (MoE)}~\cite{shazeer2017outrageously} consists of a router ${R}$ and $K$ expert networks, which in our study are implemented as LoRA modules, each denoted as $\{E_n(x) = {B}_n {A}_n x\}_{n=1}^K$.
The Router takes the input ${x}$ and computes routing scores that determine the importance of each expert.
Then, \textit{Top-k routing} is applied, activating only the $k \le K$ experts with the highest routing scores, and their scores are renormalized to yield the final routing scores, as shown in \cref{eq:Router}:
\begin{equation}
\label{eq:Router}
\hat{R}_{n} =
\begin{cases}
\dfrac{{R}_{n}(x)}
{\sum_{m \in {top}({R(x)},k)} {R}_{m}(x)}, & \text{if } n \in \text{top}({R(x)},k), \\
0, & \text{otherwise.}
\end{cases}
\end{equation}
The final output $x'$ of the MoE is obtained by taking the weighted sum of each expert’s output using the routing scores as weights, as shown in \cref{eq:moe_output}:
\begin{equation}
\label{eq:moe_output}
\mathrm{x}' = \sum_{n=1}^{K} \hat{R}_{n}(x) {E}_n({x})
\end{equation}

\subsection{Simplex Equiangular Tight Frame (ETF)}
A \textit{Simplex Equiangular Tight Frame (ETF)}~\cite{papyan2020prevalence} is a geometrically optimal structure for the perfect separation of a set of $K$ vectors ${W} = [{w}_1, \ldots, {w}_K] \in \mathbb{R}^{m \times K}$.
All its properties are captured by its ideal Gram matrix $G^{\text{ideal}} \in \mathbb{R}^{K \times K}$, as shown in \cref{eq:ideal_etf}:
\begin{equation}
\label{eq:ideal_etf}
{G}_{n,m}^\text{ideal} = {w}_n^\top {w}_m =
\begin{cases}
1, & \text{if } n = m, \\
-\frac{1}{K-1}, & \text{if } n \neq m.
\end{cases}
\end{equation}
where this condition holds for $\forall n,m\in\left[1,\ldots,K\right]$. This structure implies that all vectors have the same $\ell_2$ norm and are maximally separable (equiangular), with a pairwise cosine similarity of $-\frac{1}{K-1}$. In this study, we use an auxiliary loss function to enforce this ETF structure on our expert outputs (detailed in \cref{section:section_ETF}).
\section{Methods}
\subsection{Problem Definition}
We follow the standard ZSAD setting~\cite{zhou2023anomalyclip,qu2025bayesian,ma2025aa}, training our model via supervised learning on an auxiliary dataset of seen categories $D_s$, and then evaluating on unseen categories $D_u$ ($D_s \cap D_u = \emptyset$). During the test phase, for the $i$-th image $X_i \in \mathbb{R}^{H \times W \times 3}$, the model outputs an image-level Anomaly Score $\hat{S} \in [0, 1]$ and a pixel-level Anomaly Map $\hat{M} \in [0, 1]^{H \times W}$.
{\small
\begin{figure*}[!t]
    \centering
    \vspace{-2.5mm}
    \includegraphics[width=1\linewidth]{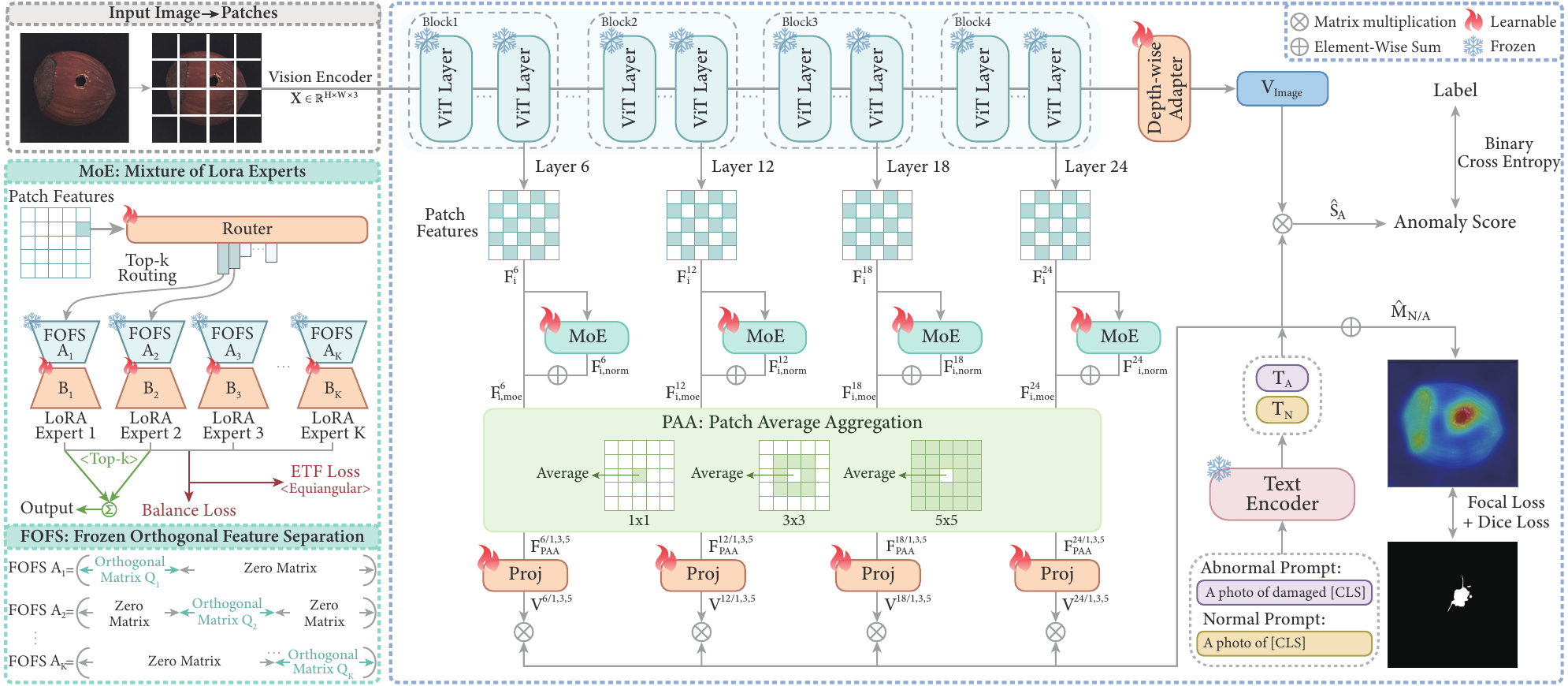}
    \vspace{-4mm}
    \caption{\textbf{The framework of MoECLIP.}
MoE is integrated into multiple layers of the CLIP Vision Encoder, enabling dynamic expert routing for each image patch to learn patch-specific representations for ZSAD. Within each MoE, FOFS enforces expert specialization by orthogonally separating the feature space and ETF loss further enhances expert diversity by maximizing the equiangular separation of expert outputs. PAA then aggregates the refined patch features across multiple scales to capture anomalies of different sizes.}
    \label{fig:main}
    \vspace{-2.5mm}
\end{figure*}
}
\subsection{Overview}
\cref{fig:main} illustrates the overall framework of MoECLIP. The core design of MoECLIP is to adapt the model for the anomaly detection task by integrating MoE modules at the output-level of multi-level layers, all while keeping the Vision Encoder weights frozen to preserve CLIP's generalization capability. Inside the MoE modules, a router dynamically selects the optimal expert based on the unique characteristics of each patch. Furthermore, to prevent functional redundancy and induce expert differentiation, we introduce two novel mechanisms: (1) Frozen Orthogonal Feature Separation (FOFS) at the LoRA input, which orthogonally separates the feature space to force experts to focus on different subspaces, and (2) a simplex Equiangular Tight Frame (ETF) loss at the LoRA output, which regulates the expert output features to be maximally equiangular.

\subsection{MoE-based Feature Adaptation}
We integrate PEFT-based Mixture of LoRA Expert modules at multiple layers, enabling each image patch to dynamically select the optimal combination of experts and adapt its representation across different levels of abstraction, all while keeping the CLIP Vision Encoder frozen. 

The MoE module at the $l$-th layer receives the corresponding patch feature $F^l_i \in \mathbb{R}^{d}$ for the $i$-th patch and learns a residual $F^l_{i, \text{expert}}\in \mathbb{R}^{d}$ to perform feature adaptation. However, a naive residual addition (i.e., $F^l_i + F^l_{i, \text{expert}}$) can cause a norm mismatch, which destabilizes training and degrades CLIP's generalization capability. Inspired by AA-CLIP~\cite{ma2025aa}, we address this by normalizing the MoE output to match the $\ell_2$ norm of the original feature while preserving its direction:
\begin{equation}
\label{eq:norm_expert_output}
F^l_{i,\text{norm}} = F^l_{i, \text{expert}} \cdot \frac{\|F^l_i\|_2}{\|F^l_{i, \text{expert}}\|_2 + \epsilon}
\end{equation}
where $\|\cdot\|_2$ denotes the $\ell_2$ norm and $\epsilon$ is a small constant for numerical stability. Subsequently, the normalized output $F^l_{i,\text{norm}}$ is combined with the original $F^l_i$ via a weighted residual connection, controlled by $\lambda_{\text{MoE}}$, to yield the patch feature $F^l_{i, \text{MoE}}$, as shown in \cref{eq:final_moe_output}:
\begin{equation}
\label{eq:final_moe_output}
F^l_{i, \text{MoE}} = \lambda_{\text{MoE}} \cdot F^l_{i,\text{norm}} + (1 - \lambda_{\text{MoE}}) \cdot F^l_i
\end{equation}

\subsection{Expert Specialization}
In this section, we introduce Frozen Orthogonal Feature Separation (FOFS) and the ETF loss to enforce expert specialization and prevent functional redundancy. We also employ a standard balance loss to prevent expert collapse, which supports this specialization.

\vspace{1ex}
\noindent\textbf{Frozen Orthogonal Feature Separation (FOFS): }
The core idea of FOFS is to separate the $d$-dimensional input feature $F^l_i \in \mathbb{R}^{d}$ into $K$ non-overlapping subspaces $c_1, \dots, c_K$, forcing each expert to focus solely on one subspace. Specifically, the LoRA matrix $A_n \in \mathbb{R}^{r \times d}$ for the $n$-th expert is defined as a block matrix. In this matrix, only the columns corresponding to the $n$-th subspace $c_n$ are filled with a random orthogonal matrix ${Q}_n \in \mathbb{R}^{r \times d_n}$ obtained via QR decomposition~\cite{gander1980algorithms}, while all other columns are zero matrices ${0}$. FOFS can be formulated as:
\begin{flalign}
& \hspace{0.8em} \begin{aligned}
    & {Q}_n = (\mathrm{qr}_{{Q}}(C_n))^T, \quad \text{where  } C_n \sim \mathcal{N}(0, \text{I})_{d_n \times r} \\
    & A_n = [{0}_{r \times d_1}, \dots, {0}_{r \times d_{n-1}}, {Q}_n, {0}_{r \times d_{n+1}}, \dots, {0}_{r \times d_K}] \raisetag{1.4\baselineskip}
\end{aligned} && \label{eq:fofs}
\end{flalign}
where $A_nA_m^T = {0}$ for all $n \neq m$, ensuring mutual orthogonality, and the subspace dimensions satisfy $\sum_{n=1}^K d_n = d$.

FOFS provides two key advantages. First, it inherently prevents redundant knowledge learning by forcing each expert to focus on a physically distinct feature subspace from initialization. Second, freezing the $A_n$ matrix helps preserve CLIP's generalization capability and reduces the risk of overfitting on auxiliary datasets. This frozen approach is inspired by recent LoRA research~\cite{zhu2024asymmetry}, which demonstrated that randomly initialized orthogonal $A$ matrices can achieve performance comparable to learned ones.

\vspace{1ex}
\noindent\textbf{ETF loss: }\label{section:section_ETF}
Even though FOFS enforces distinct roles for the experts at the input stage, their learnable LoRA $B$ matrices can still converge to similar feature spaces, leading to output stage functional redundancy. To address this, we introduce the ETF loss $\mathcal{L}_{etf}$, which enforces the $K$ expert output vectors to be maximally equiangular. Given $E^l \in \mathbb{R}^{L \times K \times d}$ representing the outputs of $K$ experts for $L$ patches at a layer $l$, we first $\ell_2$-normalize each expert vector $e^l_{i,n} \in \mathbb{R}^{d}$. Then, the Gram matrix $G_i^{l} \in \mathbb{R}^{K \times K}$ is computed for each patch $i = 1, \ldots, L$, as shown in Eq.~(\ref{eq:gram}),
\begin{equation}
\label{eq:gram}
(G^l_i)_{n,m} = (\hat{e}^l_{i,n})^\top \hat{e}^l_{i,m}
\end{equation}
where $\hat{e}_{i,n}$ and $\hat{e}_{i,m}$ denote the $\ell_2$-normalized outputs of expert $n$ and $m$ for the $i$-th patch, respectively. 
The $\mathcal{L}_{etf}$ encourages all $G_i^{l}$ to approximate the ideal equiangular Gram matrix $G^{\mathrm{ideal}}$ (\cref{eq:ideal_etf}), and is calculated as follows:
\begin{equation}
\label{eq:etf_loss}
\begin{split}
\mathcal{L}_{etf} 
&= \sum_{l} \left( \frac{1}{L} \sum_{i=1}^{L} \| G_i^{l} - G^{\mathrm{ideal}} \|_F^2 \right) \\
&= \sum_{l} \left( \frac{1}{LK^2} \sum_{i=1}^{L}\sum_{n=1}^{K}\sum_{m=1}^{K} \big((G^l_i)_{n,m} - G^{\mathrm{ideal}}_{n,m}\big)^2 \right)
\end{split}
\end{equation}
By enforcing expert outputs to span diverse and equiangular directions at each stage, $\mathcal{L}_{etf}$ complements FOFS by further boosting expert specialization at the output stage.

\vspace{1ex}
\noindent\textbf{Balance loss: }
To prevent expert collapse and ensure assignments are distributed evenly, we use the standard balance loss~\cite{shazeer2017outrageously,meng2024moead} $\mathcal{L}_{bal}$, as shown in \cref{eq:balance_loss},
\begin{equation}
\label{eq:balance_loss}
\mathcal{L}_{bal} = \sum_{l} \left[ \text{CV}^2\left(\sum_{i=1}^{L} R(F^l_i)\right) \right]
\end{equation}
where $L$ is the number of patches, $R(F^l_i) \in \mathbb{R}^K$ is the routing probability vector and $\text{CV}^2(\cdot)$ denotes the squared Coefficient of Variation (detailed in supplementary \cref{sec:balance}).

\subsection{Patch Average Aggregation (PAA)}
The CLIP Vision Encoder's (ViT) fixed-size patch division limits its ability to effectively detect anomalies of varying scales. To our knowledge, while previous works~\cite{li2024musc,pan2025pa} use patch aggregation to leverage contextual information, these methods are limited to only the test phase, thus failing to integrate multi-scale awareness during the training phase. We therefore apply the parameter-free Patch Average Aggregation (PAA) module during the training phase to the refined patch features $F^l_{\text{MoE}} \in \mathbb{R}^{{L} \times d}$. By enabling each patch to leverage neighboring context, PAA incorporates multi-scale awareness to integrate fragmented anomaly patterns across boundaries, enhancing structural robustness.

To apply PAA, we first reshape the patch embeddings $F^l_{\text{MoE}}$ into a 2D spatial grid $P^l \in \mathbb{R}^{\sqrt{L} \times \sqrt{L} \times d}$. A new patch feature $\hat{p}^l_{h,w}$ is generated by calculating the average of all patch features within an $s \times s$ sliding window centered at coordinate $(h, w)$, as shown in \cref{eq:paa},
\begin{equation}
\label{eq:paa}
\hat{p}^l_{h,w} = \frac{1}{s^2} \sum_{u=-\lfloor s/2 \rfloor}^{\lfloor s/2 \rfloor} \sum_{v=-\lfloor s/2 \rfloor}^{\lfloor s/2 \rfloor} p^l_{h+u, w+v}
\end{equation}
where $p^l_{h+u, w+v}$ denotes the original patch feature at coordinate $(h+u, w+v)$, and $s$ is a positive odd integer representing the window size. This aggregation process is performed independently for each scale $s$, extracting multiple sets of patch features $\hat{P}^{l,s}$. Finally, $\hat{P}^{l,s}$ is reshaped back into $\hat{F}^{l,s}_{PAA} \in \mathbb{R}^{L \times d}$ to form the multi-scale patch features.

\subsection{Anomaly Map and Anomaly Score}
To compute the final outputs, we first prepare the representative text features $T_A$ and $T_N$ from text prompts. Following AA-CLIP~\cite{ma2025aa}, we pass a set of text prompts (detailed in supplementary \cref{sec:prompt}) through the pretrained CLIP Text Encoder and obtain $T_A, T_N \in \mathbb{R}^d$ by averaging the outputs of their corresponding text prompt sets. These features are subsequently used to calculate the pixel-level Anomaly Map and the image-level Anomaly Score.

\vspace{1ex}
\noindent\textbf{Anomaly Map: } 
First, the PAA features $\hat{F}^{l,s}_{\mathrm{PAA}}$ are aligned with the text space via a trainable projection layer $\mathrm{Proj}_l$, yielding $V^{l,s} \in \mathbb{R}^{L \times d}$, as shown in Eq.~(\ref{eq:proj_layer}):
\begin{equation}
\label{eq:proj_layer}
V^{l,s} = Proj_l(\hat{F}^{l,s}_{{PAA}})
\end{equation}
This text-aligned feature $V^{l,s}$ is then used to calculate the anomaly map $\hat{M}_{N/A}^{l,s} \in \mathbb{R}^{H \times W \times 2}$ for the $l$-th layer and $s$-th scale, as shown in Eq.~(\ref{eq:anomaly_map_lr}),
\begin{equation}
\label{eq:anomaly_map_lr}
\hat{M}_{N/A}^{l,s} = \text{softmax}(\phi(\mathrm{cos}(V^{l,s}, T_{{N/A}})))
\end{equation}
where $\phi$ denotes an interpolation function that resizes the map to the original image dimensions, and $\mathrm{cos}(\cdot, \cdot)$ represents the cosine similarity.

\vspace{1ex}
\noindent\textbf{Anomaly Score: }
The image-level anomaly score is determined by the semantic alignment between the patch features $\hat{F}^{f}_{\mathrm{PAA}}$ from the final layer and the text features. To enhance this alignment, $\hat{F}^{f}_{\mathrm{PAA}}$ passes through a Depth-wise Adapter. This adapter, inspired by MobileNet~\cite{howard2017mobilenets}, uses a 1D depthwise separable convolution, which combines a Depth-wise and a Point-wise convolution to efficiently capture features with fewer parameters, making it lightweight and less prone to overfitting. The resulting features are then aggregated into $V_{\mathrm{image}} \in \mathbb{R}^d$ via Global Average Pooling. The Depth-wise Adapter can be calculated as:
\begin{equation}
\begin{aligned}
    &\hat{F}_{\mathrm{dw}} = GELU({DwConv1d}({LN}(\hat{F}^{f}_{\mathrm{PAA}}))) \\
    &V_{\mathrm{image}} = \frac{1}{L} \sum_{i=1}^{L} {PwConv1d}(\hat{F}_{\mathrm{dw}})_i
\end{aligned}
\end{equation}
where ${LN}(\cdot)$ is a Layer Normalization. The $V_{\mathrm{image}}$ is compared with the text embeddings $T_A$ and $T_N$ via cosine similarity, yielding the anomaly score vector 
$\hat{S}_{N/A} \in \mathbb{R}^2$, as shown in Eq.~(\ref{eq:anomaly_score}),
\begin{equation}
\label{eq:anomaly_score}
\hat{S}_{N/A} =
\mathrm{cos}\left(V_{\mathrm{image}}, [T_A, T_N]\right)
\end{equation}
where $[\,\cdot\,,\,\cdot\,]$ denotes the concatenation operation, and $\hat{S}_A$ is finally used as the image-level anomaly score.

\subsection{Loss Functions}
Following previous works~\cite{chen2023april,zhou2023anomalyclip,cao2024adaclip,gu2024filo,ma2025aa}, we employ a combination of Focal~\cite{focalloss} and Dice Loss~\cite{diceloss} for the anomaly segmentation loss $\mathcal{L}_{seg}$, and the Binary Cross-Entropy loss for the anomaly classification objective $\mathcal{L}_{ac}$, as shown in Eq.~(\ref{eq:loss_ac_seg}),
\begin{equation}
\label{eq:loss_ac_seg}
\begin{aligned}
    \mathcal{L}_{seg} &= \sum_{l} \sum_{s} \Big[ \text{Focal}(\hat{M}^{l,s}_{N/A}, M) + \text{Dice}(\hat{M}^{l,s}_{N/A}, M)\Big] \\
    \mathcal{L}_{ac} &= \text{BCE}(\hat{S}_A, S)
\end{aligned}
\end{equation}
where $M \in \mathbb{R}^{H \times W}$ denotes the pixel-level ground-truth segmentation mask, and ${S}$ denotes the image-level label indicating whether the image is anomalous.

The final loss function can be expressed as:
\begin{equation}
\label{eq:total_loss}
\mathcal{L}_{total} = \mathcal{L}_{seg} + \mathcal{L}_{ac} + \lambda_{etf} \mathcal{L}_{etf} + \lambda_{bal} \mathcal{L}_{bal}
\end{equation}
where $\mathcal{L}_{etf}$ and $\mathcal{L}_{bal}$ are the auxiliary losses for expert specialization, calculated as in Eq.~(\ref{eq:etf_loss}) and Eq.~(\ref{eq:balance_loss}) respectively.
\begin{table*}[htbp]
    \centering

    \renewcommand{\arraystretch}{1.2}
    \caption{Comparison with state-of-the-art methods across industrial and medical domains under the ZSAD setting. The symbol † indicates results obtained from models re-trained under our setting. The best performance is in \textbf{bold} and the second-best is \underline{underlined}.}
    \vspace{-2mm}
    \resizebox{\textwidth}{!}{%

    \begin{tabular}{cccccccccc}
        \toprule

        metric & Domain & Dataset & \begin{tabular}[c]{@{}c@{}}WinCLIP \cite{jeong2023winclip} \\ (CVPR 2023)\end{tabular} & \begin{tabular}[c]{@{}c@{}}April-GAN \cite{chen2023april} \\ (CVPRw 2023)\end{tabular} & \begin{tabular}[c]{@{}c@{}}AnomalyCLIP \cite{zhou2023anomalyclip} \\ (ICLR 2024)\end{tabular} & \begin{tabular}[c]{@{}c@{}}†AdaCLIP \cite{cao2024adaclip} \\ (ECCV 2024)\end{tabular} & \begin{tabular}[c]{@{}c@{}}†AA-CLIP \cite{ma2025aa} \\ (CVPR 2025)\end{tabular} & \begin{tabular}[c]{@{}c@{}}Bayes-PFL \cite{qu2025bayesian} \\ (CVPR 2025)\end{tabular} & \begin{tabular}[c]{@{}c@{}}\textbf{MoECLIP} \\ \textbf{(Ours)}\end{tabular} \\
        \midrule

        % --- Image-level Section ---

        \multirow{9}{*}{\begin{tabular}[c]{@{}c@{}}Image-level \\ (AUROC, AP)\end{tabular}} & 

        \multirow{5}{*}{Industrial} & MVTec-AD & (91.8, 95.1) & (86.1, 93.6) & (91.9, 96.2) & (91.5, \underline{96.1}) & (90.9, 96.0) & (\underline{92.2}, \underline{96.1}) & (\textbf{93.9}, \textbf{96.8}) \\
        & & VisA & (78.1, 77.5) & (77.5, 80.8) & (82.1, 85.4) & (\underline{83.0}, 85.5) & (79.2, 83.7) & (\textbf{86.8}, \textbf{89.3}) & (\underline{83.6}, \underline{86.2})\\
        & & BTAD & (83.3, 84.1) & (73.4, 69.6) & (92.5, 94.2) & (91.6, 92.4) & (\textbf{94.8}, \underline{97.5}) & (93.0, 96.7) & (\underline{93.1}, \textbf{98.0}) \\
        & & RSDD & (85.3, 65.3) & (72.7, 68.3) & (74.0, 73.2) & (83.8, 80.1) & (\underline{94.9}, \underline{94.2}) & (91.3, 89.7) & (\textbf{95.3}, \textbf{95.1}) \\
        & & DTD-Synthetic & (\underline{95.0}, \underline{97.9}) & (85.5, 94.0) & (93.3, 97.7) & (91.5, 96.3) & (92.5, 97.7) & (93.5, 97.7) & (\textbf{95.5, 98.6}) \\
        \cline{2-10}
        & 

        \multirow{4}{*}{Medical} & Brain MRI & (45.1, 80.3) & (58.8, 87.8) & (70.8, 90.6) & (79.5, \underline{94.5}) & (79.6, 94.4) & (\underline{81.9}, \underline{94.5}) & (\textbf{88.5, 97.1}) \\
        & & Head CT & (83.7, 81.6) & (86.9, 87.9) & (95.1, 95.3) & (\underline{95.7}, 93.2) & (95.4, \underline{94.3}) & (95.4, 93.2) & (\textbf{96.6}, \textbf{94.5}) \\
        & & Liver CT & (66.5, 56.1) & (54.7, 49.6) & (\underline{68.2}, \underline{63.4}) & (62.6, 54.2) & (58.4, 49.7) & (61.7, 55.2) & (\textbf{74.0}, \textbf{64.6}) \\
        & & Retina OCT & (53.7, 44.3) & (65.6, 60.5) & (74.7, 73.9) & (70.3, 69.4) & (83.4, \underline{83.8}) & (\underline{83.7}, 81.8) & (\textbf{85.5}, \textbf{84.9}) \\
        
        \cline{2-10}

        % --- Image-level Average Row ---

        & \multicolumn{2}{c}{\begin{tabular}[c]{@{}c@{}}Average\end{tabular}} & \begin{tabular}[c]{@{}c@{}}(75.8, 75.8)\end{tabular} & \begin{tabular}[c]{@{}c@{}}(73.5, 76.9)\end{tabular} & \begin{tabular}[c]{@{}c@{}}(82.5, 85.5)\end{tabular} & \begin{tabular}[c]{@{}c@{}}(83.3, 84.6)\end{tabular} & \begin{tabular}[c]{@{}c@{}}(85.5, 87.9)\end{tabular} & \begin{tabular}[c]{@{}c@{}}(\underline{86.6}, \underline{88.2})\end{tabular} &
        \begin{tabular}[c]{@{}c@{}}(\textbf{89.6}, \textbf{90.6})\end{tabular} \\

        \midrule

        % --- Pixel-level Section ---

        \multirow{13}{*}{\begin{tabular}[c]{@{}c@{}}Pixel-level \\ (AUROC, AP)\end{tabular}} & 

        \multirow{5}{*}{Industrial} & MVTec-AD & (85.1, 18.0) & (87.6, 40.8) & (88.1, 38.6) & (89.7, 43.0) & (91.6, 45.4) & (\underline{91.9}, \textbf{48.4}) & (\textbf{92.5}, \underline{45.7}) \\
        & & VisA & (79.6, 5.0) & (94.2, 25.8) & (\underline{95.5}, 21.2) & (\underline{95.5}, \underline{28.6}) & (94.7, 24.2) & (\underline{95.5}, \textbf{29.2}) & (\textbf{95.6}, 26.1)\\
        & & BTAD & (71.4, 11.2) & (91.4, 32.4) & (94.2, 42.0) & (95.4, 47.0) & (\underline{95.6}, 49.4) & (\underline{95.6}, 48.6) & (\textbf{96.8}, \textbf{50.4}) \\
        & & RSDD & (95.1, 2.1) & (99.3, 33.1) & (98.6, 17.8) & (\underline{99.6}, 33.2) & (99.4, \textbf{41.7}) & (\underline{99.6}, 35.7) & (\textbf{99.7}, \underline{35.9}) \\
        & & DTD-Synthetic & (82.5, 11.6) & (96.5, 67.7) & (96.2, 54.0) & (96.4, 57.7) & (97.6, 61.7) & (\underline{98.2}, \textbf{66.7}) & (\textbf{98.8}, \underline{62.7}) \\
        \cline{2-10} 
        & 

        \multirow{8}{*}{Medical} & Brain MRI & (95.4, 23.5) & (94.4, 37.1) & (96.0, 57.5) & (94.5, 35.6) & (\underline{96.7}, \underline{55.1}) & (95.7, 42.9) & (\textbf{97.3}, \textbf{61.3}) \\
        & & Liver CT & (\underline{97.1}, 8.0) & (95.5, 5.3) & (93.0, 2.9) & (97.0, \underline{9.4}) & (\textbf{97.2}, 9.3) & (96.5, 6.2) & (\textbf{97.2}, \textbf{10.8}) \\
        & & Retina OCT & (88.8, 22.0) & (88.6, 35.4) & (91.8, 47.1) & (94.4, 55.5) & (95.4, \underline{62.3}) & (\underline{95.5}, 55.0) & (\textbf{96.2}, \textbf{66.3}) \\
        & & ColonDB & (64.8, 14.3) & (78.3, 23.3) & (81.6, 34.0) & (81.0, 26.5) & (82.8, \underline{31.5}) & (\underline{82.9}, 30.7) & (\textbf{85.4}, \textbf{34.8}) \\
        & & ClinicDB & (70.7, 19.4) & (85.0, 38.5) & (84.3, 41.1) & (85.7, 45.8) & (\underline{89.2}, \underline{49.8}) & (88.2, 49.1) & (\textbf{89.7}, \textbf{49.9}) \\
        & & CVC-300 & (44.0, 5.0) & (92.9, 26.8) & (95.8, 56.4) & (91.1, 25.9) & (96.5, \textbf{53.9}) & (\underline{96.6}, 51.6) & (\textbf{97.0}, \underline{53.0}) \\
        & & Endo & (68.2, 23.8) & (84.0, 48.8) & (85.4, 49.7) & (86.2, 54.1) & (88.6, 57.7) & (\underline{89.4}, \underline{58.9}) & (\textbf{91.0}, \textbf{62.5}) \\
        & & Kvasir & (69.8, 27.5) & (79.7, 43.0) & (81.4, 43.1) & (82.9, 51.3) & (\underline{86.0}, 52.9) & (85.6, \underline{53.4}) & (\textbf{88.1}, \textbf{57.6}) \\
        
        \cline{2-10} 
        
        % --- Pixel-level Average Row ---

        & \multicolumn{2}{c}{\begin{tabular}[c]{@{}c@{}}Average\end{tabular}} & \begin{tabular}[c]{@{}c@{}}(77.9, 14.7)\end{tabular} & \begin{tabular}[c]{@{}c@{}}(89.8, 35.2)\end{tabular} & \begin{tabular}[c]{@{}c@{}}(90.9, 38.9)\end{tabular} & \begin{tabular}[c]{@{}c@{}}(91.5, 39.5)\end{tabular} & \begin{tabular}[c]{@{}c@{}}(\underline{93.2}, \underline{45.8})\end{tabular} & \begin{tabular}[c]{@{}c@{}}(\underline{93.2}, 44.3)\end{tabular} & \begin{tabular}[c]{@{}c@{}}(\textbf{94.3}, \textbf{47.5})\end{tabular} \\

        \bottomrule
    \end{tabular}}
\label{table:performance}
\end{table*}

\section{Experiments}
\subsection{Experimental Setup}
\noindent\textbf{Datasets: }We evaluate our model's performance on a comprehensive suite of 14 datasets, spanning 5 industrial and 9 medical datasets. The industrial datasets include MVTec-AD~\cite{mvctec}, VisA~\cite{visa}, BTAD~\cite{BTAD}, RSDD~\cite{RSDD}, and DTD-Synthetic~\cite{DTD-Synthetic}. The medical datasets cover various tasks, including Head CT~\cite{HeadCT} for brain tumor detection, three datasets from the BMAD benchmarks~\cite{bmad} (BrainMRI, Liver CT, and Retina OCT), and five datasets for colon polyp detection (CVC-ColonDB~\cite{ColonDB}, CVC-ClinicDB~\cite{ClinicDB}, CVC-300~\cite{cvc300}, Endo~\cite{Endo}, and Kvasir~\cite{kvasir}). We use VisA as the training dataset for all evaluations on other datasets. To ensure a fair comparison, VisA~\cite{visa} results are obtained using a model trained on MVTec-AD~\cite{mvctec}. A detailed description of the datasets is provided in supplementary \cref{sec:data}.

\vspace{1ex}
\noindent\textbf{Evaluation Metrics: }We evaluate ZSAD performance on both image-level classification and pixel-level segmentation using the Area Under the Receiver Operating Characteristic (AUROC) and Average Precision (AP).

\vspace{1ex}
\noindent\textbf{Implementation details: }We utilize the OpenCLIP ViT-L/14-336 architecture pre-trained by OpenAI~\cite{radford2021learning} as our backbone. All parameters of the pretrained CLIP model are kept frozen during training, and all input images are resized to $518 \times 518$. Our MoE modules are integrated at the output of the $l$-th layers of the Vision Encoder, where $l \in \{6, 12, 18, 24\}$, and their outputs are used to compute the final outputs. For the model configuration, we set the total number of experts $K$ to 4 with Top-2 routing via a linear router layer ${R}$, a LoRA rank $r$ to 8, MoE residual weight $\lambda_{\mathrm{MoE}}$ to 0.1, PAA scales $s \in \{1, 3, 5\}$, the auxiliary loss weights $\lambda_{etf}$ and $\lambda_{bal}$ to 0.01. The model is trained for 20 epochs using the Adam Optimizer with a learning rate of $5 \times 10^{-4}$. All experiments are conducted on 2 × NVIDIA Tesla V100 16GB GPUs. More details can be found in supplementary \cref{sec:setup}.

{\small
\begin{figure*}[t]
    \centering
    \vspace{-2.5mm}
    \includegraphics[width=1\linewidth]{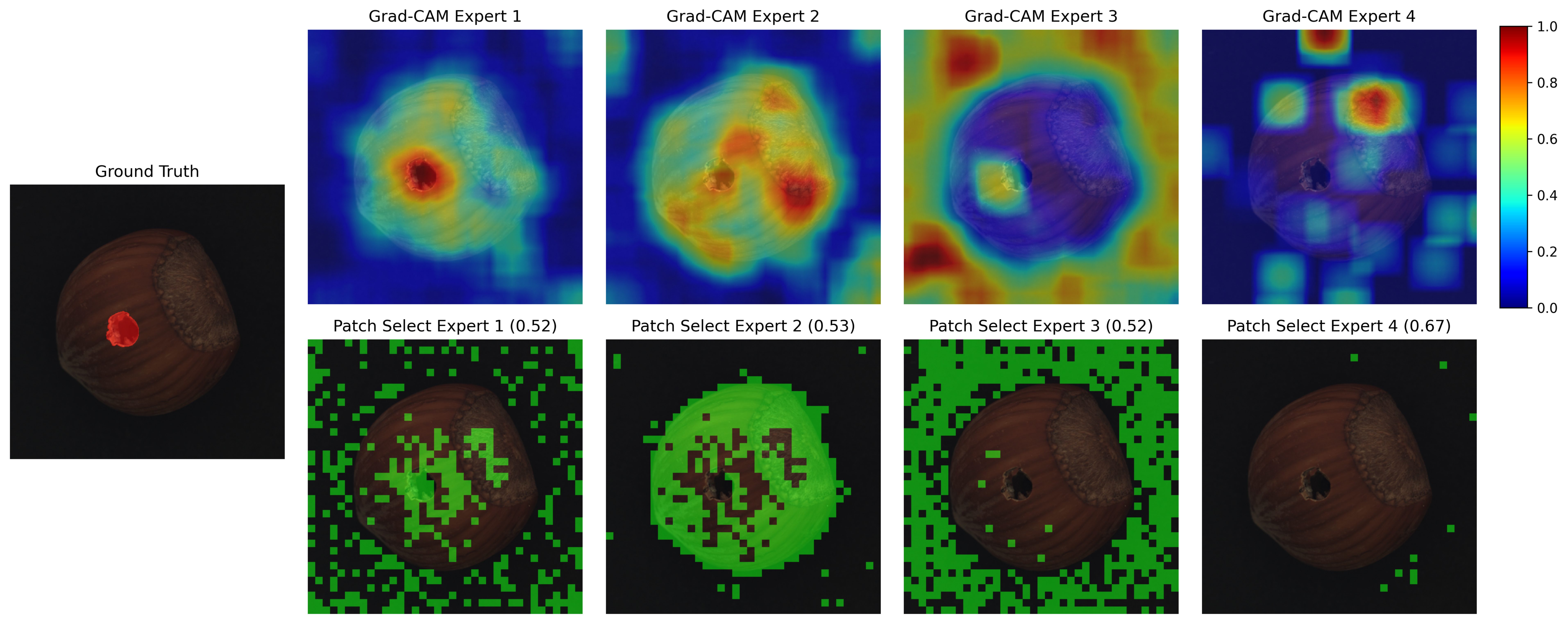}
    \vspace{-4mm}
    \caption{\textbf{Visualization of Grad-CAM and Patch Selection Map for each expert at layer 18.}
The Ground Truth image is shown on the far left. The top row (Grad-CAM) highlights each expert's focus region. The bottom row (Patch Selection) illustrates the patches where the corresponding expert was the router's Top-1 choice (shown in green). The value in each subplot title represents the expert's average renormalized routing weight based on the Top-1 setting for its Top-1 assigned patches.}
    \label{fig:gradcam}
    \vspace{-2.5mm}
\end{figure*}
}

\begin{figure}[tb]
    \centering

    \makebox[\columnwidth][c]{%
        \hspace*{-0.01\columnwidth}
        \includegraphics[width=1.05\columnwidth, keepaspectratio]{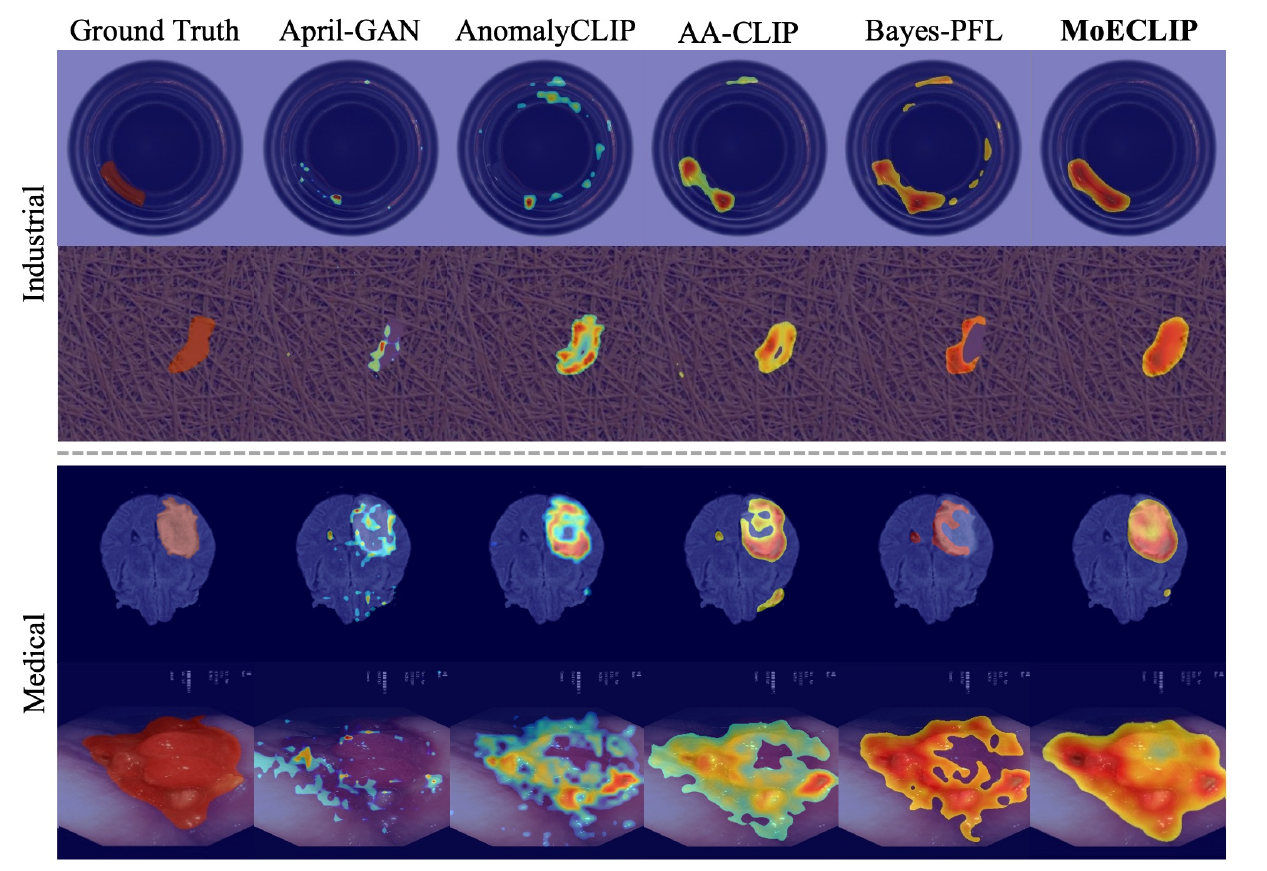}
    }

    \setlength{\abovecaptionskip}{6pt}   
    \setlength{\belowcaptionskip}{2pt}
    % --------------------------------

    \caption{\textbf{Visualization of Anomaly Maps comparing MoECLIP with previous ZSAD methods across industrial and medical domains.} 
    The first column shows the Ground Truth, and the remaining columns show anomaly maps from each method.}
    \label{fig:anomalymap}
    \vspace{-1.5mm}
\end{figure}

\subsection{Comparison with State-of-the-art methods}
\cref{table:performance} presents a comprehensive ZSAD performance comparison between MoECLIP and recent state-of-the-art (SOTA) methods, including WinCLIP~\cite{jeong2023winclip}, April-GAN~\cite{chen2023april}, AnomalyCLIP~\cite{zhou2023anomalyclip}, AdaCLIP~\cite{cao2024adaclip},  AA-CLIP~\cite{ma2025aa}, and Bayes-PFL~\cite{qu2025bayesian}. For fairness, all competing models are trained on VisA as the auxiliary dataset under a unified setting (detailed in supplementary \cref{sec:method}).

Overall, our MoECLIP consistently achieves SOTA performance across both domains. Specifically, for image-level metrics, it further yields improvements of 3.0\(\%\) in AUROC and 2.4\(\%\) in AP. For pixel-level metrics, MoECLIP outperforms the second-best method by 1.1\(\%\) in average AUROC and 1.7\(\%\) in average AP. These results confirm that our patch-level dynamic routing is substantially more effective for ZSAD than uniform adaptation strategies employed in previous works. Furthermore, the strong performance on medical datasets shows that patch-specialized experts, despite being trained only on industrial data, transfer and generalize robustly to distinct medical domains.

\subsection{Visualization}
\cref{fig:gradcam} visualizes the behavior of the 18th-layer experts on the MVTec-AD hazelnut class to assess whether MoECLIP performs patch-specialized routing. We analyze both the experts’ focus regions (top row, Grad-CAM~\cite{selvaraju2017grad}) and the router’s decisions (bottom row, patch selection maps), which show the Top-1 selected expert for visual clarity. The visualization demonstrates a clear functional differentiation. The Grad-CAM reveals that the experts have learned to focus on distinct regions of the image: Expert 1 focuses on the anomaly, Expert 2 on the object body, Expert 3 on the background, and Expert 4 is rarely utilized (detailed in supplementary~\cref{sec:Utilization}). The Patch Selection maps show that the router allocates patches in alignment with these focus patterns: Expert 1 is the Top-1 choice for anomaly-related and some background patches, whereas Experts 2 and 3 mainly receive object-body and background patches, respectively. Furthermore, the average routing weights show comparable values across Expert 1, 2, and 3. Expert 4's weight is unrepresentative, being rarely selected. This confirms the anomaly is identified by dynamic combination of functionally distinct, patch-specialized experts, not a single one, demonstrating content-based routing. More visualizations for the other layers and the Top-2 selected experts are provided in supplementary \cref{sec:gradcam}.

\cref{fig:anomalymap} provides a comparison of Anomaly Maps between MoECLIP and existing SOTA methods on both industrial and medical domains. Due to its patch-specialized expert design, MoECLIP delivers more accurate and fine-grained anomaly localization than existing methods.

\begin{table*}[htbp]
    \centering
    \renewcommand{\arraystretch}{0.8}
    \footnotesize
    \caption{Ablation study on the different components of MoECLIP, evaluated on industrial datasets MVTec-AD and DTD-Synthetic, and medical datasets Head CT and ColonDB using (Pixel-level AUROC, Image-level AUROC) metrics. The best performance is in \textbf{bold}.}
    \vspace{-2mm}
    \begin{tabular*}{\textwidth}{ll|@{}>{\hspace{10pt}}c@{\extracolsep{\fill}}>{\hspace{10pt}}c>{\hspace{10pt}}c>{\hspace{10pt}}c>{\hspace{10pt}}c}
        \toprule
        \multicolumn{2}{c|}{\multirow{2}{*}[-1.5ex]{Components of MoECLIP}} & \multicolumn{5}{c}{Datasets} \\[2pt]
        \cline{3-7}
        \multicolumn{2}{c|}{} 
        & \rule{0pt}{10pt}MVTec-AD 
        & DTD-Synthetic 
        & Head CT & ColonDB & Average \\
        \midrule
        \multicolumn{1}{c}{base} & \multicolumn{1}{c|}{Vanilla CLIP} & (38.4, 74.1) & (33.9, 71.6) & ($-$, 56.5) & (49.5, $-$) & (40.6, 67.4) \\
        \midrule
        (a) & w/o FOFS \& ETF Loss & (91.6, 91.7) & (97.8, 93.1) & ($-$, 94.4) & (84.1, $-$) & (91.2, 93.1) \\
        (b) & w/o FOFS & (92.0, 92.8) & (98.3, 93.9) & ($-$, 95.0) & (85.3, $-$) & (91.9, 93.9) \\
        (c) & w/o ETF Loss & (92.2, 92.7) & (98.2, 93.4) & ($-$, 96.1) & (84.6, $-$) & (91.7, 94.1) \\
        (d) & w/o Depth-wise Adapter & (92.0, 92.5) & (98.1, 93.8) & ($-$, 94.5) & (85.0, $-$) & (91.7, 93.6) \\
        (e) & w/o PAA & (92.1, 92.8) & (98.1, 94.7) & ($-$, 93.1) & (81.9, $-$) & (90.7, 93.5) \\
        \midrule
        \multicolumn{1}{c}{Ours} & \multicolumn{1}{c|}{\textbf{MoECLIP}} & \textbf{(92.5, 93.9)} & \textbf{(98.8, 95.5)} & \textbf{($-$, 96.6)} & \textbf{(85.4, $-$)} & \textbf{(92.2, 95.3)} \\
        \bottomrule
    \end{tabular*}
    \label{table:ablation}
\end{table*}

\begin{figure}[tb]
    \centering
    \includegraphics[width=1.0\columnwidth]{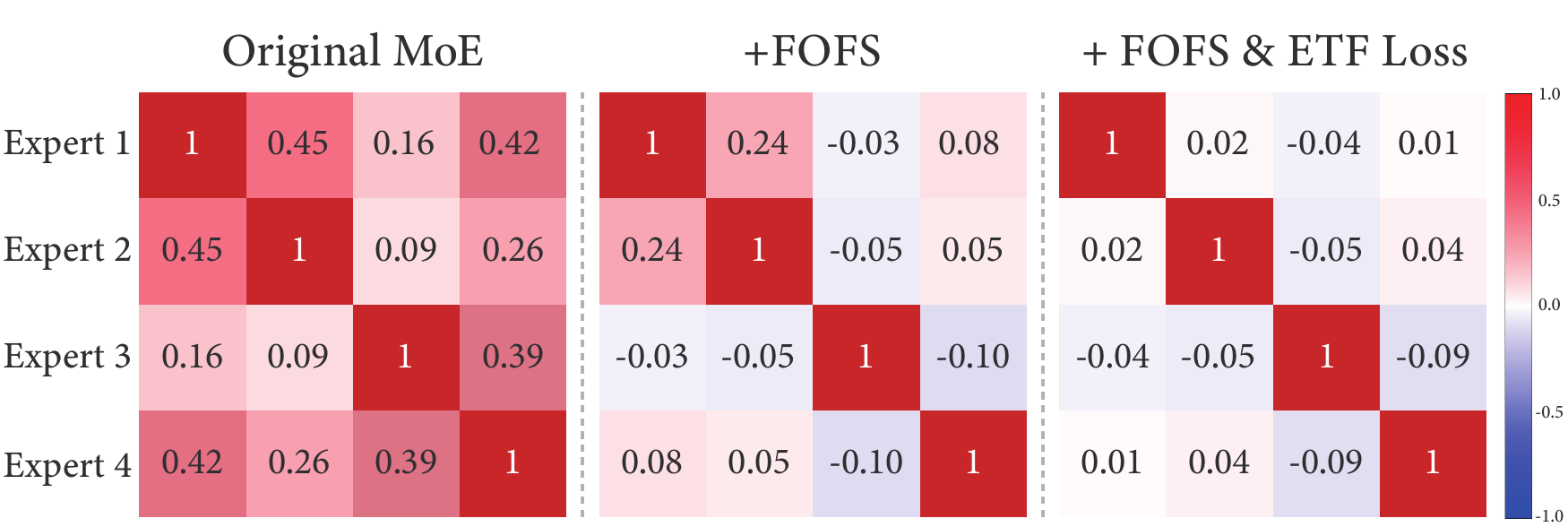}
    \caption{\textbf{Ablation Study on Inter-Expert similarity heatmap at layer 18.} The heatmap shows the average pairwise cosine similarity between expert features, computed on the MVTec-AD test set. Values approaching +1 (red) indicate high redundancy, while values approaching 0 (white) or negative values (blue) signify successful differentiation.}
    \label{similarity}
\vspace{-1.5mm}
\end{figure}

\subsection{Ablation Study}
\vspace{1ex}
\noindent\textbf{Impact of components: }
\cref{table:ablation} presents ablation results on the 2 industrial datasets and 2 medical datasets to evaluate the impact of the key components in MoECLIP. Specifically, (a) Compared to the Vanilla CLIP baseline, the model lacking specialization still achieves a significant performance improvement. However, this model performs worse than our MoECLIP model as it suffers from functional redundancy (\cref{similarity}), which our proposed FOFS and ETF Loss solve to achieve the final performance.
(b-c) Removing either the FOFS strategy or the ETF Loss individually causes a performance drop, confirming both components are complementary and essential for resolving the functional redundancy mentioned in (a).
(d) Removing the Depth-wise Adapter degrades both image-level and pixel-level performance, indicating its effectiveness in refining features.
(e) Notably, removing the PAA module causes a significant performance decrease on medical datasets. This highlights that aggregating multi-scale contextual information via PAA is important for medical domains.

\vspace{1ex}
\noindent\textbf{Functional Redundancy: }
To quantitatively assess functional redundancy, we analyze the inter-expert cosine similarity as shown in \cref{similarity}. This matrix is computed by averaging the similarity between expert-specific features, which are derived by averaging all patch features that selected that expert (via Top-k) within each image, across the MVTec-AD dataset.
The Original MoE exhibits high similarity, with the value between Expert 1 and 2 reaching $0.45$, indicating significant functional redundancy. While introducing the FOFS strategy alone substantially reduces this similarity, a notable positive similarity of $0.24$ persists between Expert 1 and 2. Finally, the full MoECLIP model, combining FOFS with the ETF Loss, minimizes all remaining similarity, bringing the value between Expert 1 and 2 down to $0.02$. This demonstrates that our approach forces experts to learn specialized, non-overlapping functions.

\begin{figure}[tb]
    \centering
    \includegraphics[width=1.0\columnwidth]{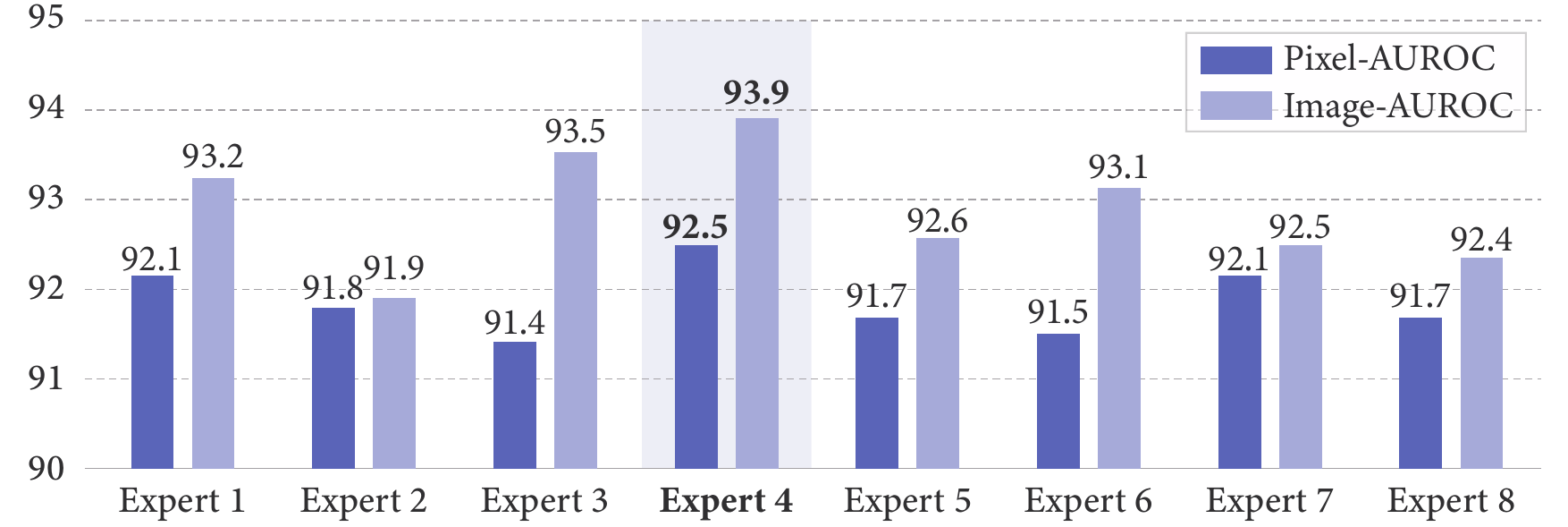}
    \caption{\textbf{Ablation Study on the Number of Experts $K$ (from 1 to 8) on the MVTec-AD dataset.}}
    \label{number_expert}
\vspace{-1.5mm}
\end{figure}

\vspace{1ex}
\noindent\textbf{Impact of the Number of Experts: }
As shown in \cref{number_expert}, increasing the number of experts does not necessarily lead to performance improvements in the ZSAD task. We empirically set $K=4$ as it yields the best results across both pixel-level and image-level AUROC. This suggests a trade-off: too few experts may under-specialize, while too many can cause functional redundancy.

\vspace{1ex}
\noindent{Additional analyses are provided in supplementary \cref{sec:experiment}}

\section{Conclusion}
In this paper, we propose MoECLIP, a novel MoE-based framework that introduces patch-level adaptation to the ZSAD task, overcoming the patch-agnostic design limitations of existing approaches. MoECLIP achieves this by dynamically routing each image patch to a specialized LoRA expert based on its unique characteristics.
Furthermore, to solve the inherent functional redundancy problem in MoE and force expert differentiation, FOFS orthogonally separates the input feature space, and an ETF loss enforces an ideal equiangular structure on the expert outputs. With this design, MoECLIP demonstrates its effectiveness and practical utility through comprehensive experiments on benchmark datasets spanning industrial and medical domains, consistently outperforming existing SOTA methods.

\section*{Acknowledgement}
This research was supported by the Korea Institute for Advancement of Technology (KIAT) grant funded by the Korea Government (MOTIE) (No. P0023675, HRD Program for Industrial Innovation).

{
    \small
    \bibliographystyle{ieeenat_fullname}
    \bibliography{main}
}
\clearpage
\raggedbottom
\providecommand{\theHsection}{\thesection}
\renewcommand{\theHsection}{Supp.\thesection}

\providecommand{\theHequation}{\theequation}
\renewcommand{\theHequation}{Supp.\theequation}
\maketitlesupplementary
\algrenewcommand{\algorithmiccomment}[1]{\(\triangleright\) #1}

\makeatletter
\@addtoreset{equation}{section}
\makeatother
\setcounter{section}{0}
\renewcommand{\thesection}{\Alph{section}}
\renewcommand{\theequation}{\thesection.\arabic{equation}}
\startcontents[supplementary]
\begingroup
    \setlength{\parskip}{0pt plus 1fill}
    \renewcommand{\numberline}[1]{#1 \hspace{0.75em}}
    \printcontents[supplementary]{l}{1}{\section*{Contents}}
    \vfill 
\endgroup
\newpage

\section{Algorithm Implementation}
\label{sec:Algorithm}
\cref{algorithm:moeclip_alg} provides pseudocode to illustrate MoECLIP's core feature adaptation logic with specialization.

\begin{algorithm}[H]
\caption{MoE Feature Adaptation with Specialization}
\label{algorithm:moeclip_alg}

\textbf{Input:} Patch features $F^l \in \mathbb{R}^{L \times d}$ from the $l$-th ViT layer; 
MoE module $M^l$ with $K$ experts; Top-$k$ value $k$. \\[3pt]
\textbf{Output:} Adapted patch features $F^l_{\text{MoE}}$; ETF Loss $\mathcal{L}_{etf}$.

\begin{algorithmic}[1]
\vspace{6pt}

%---------------------------------------------
\Statex \Comment{\textbf{Step 1:} FOFS-based expert initialization}
\State Initialize experts $\{ E_n^l(A_n^l, B_n^l) \}_{n=1}^K$ using FOFS (\cref{eq:fofs}); 
$A_n^l$ frozen, $B_n^l$ trainable.
\vspace{6pt}

%---------------------------------------------
\Statex \Comment{\textbf{Step 2:} Routing scores for all patches}
\State $R^l \leftarrow \mathrm{Softmax}(\mathrm{Router}^l(F^l))$
\hfill ($R^l \in \mathbb{R}^{L \times K}$)
\vspace{8pt}

%---------------------------------------------
\Statex \Comment{\textbf{Step 3:} Perform patch-wise routing and adaptation}
\State $\mathcal{E} \leftarrow [\ \ ]$ \Comment{Initialize list to collect expert outputs for ETF loss}
\For{each patch $i = 1,\dots,L$}
    \State Extract patch routing scores $R^l_{i}$ from $R^l$.
    \State Compute normalized Top-k weights $\hat{R}^l_{i}$ (\cref{eq:Router}).
    
    \State \Comment{Compute all K expert outputs}
    \State $\mathcal{E}_i \leftarrow [E_n^l(F^l_i) \text{ for } n=1,\dots,K]$
    \State $\mathcal{E}.\text{append}(\mathcal{E}_i)$
    
    \State \Comment{Compute weighted sum using normalized weights}
    \State $F^l_{i,\text{expert}} \leftarrow \sum_{n=1}^K \hat{R}^l_{i,n} \cdot \mathcal{E}_{i, n}$
    
    \State \Comment{Normalization and residual connection}
    \State $F^l_{i,\text{norm}} \leftarrow F^l_{i,\text{expert}} \cdot (||F^l_i||_{2} / (||F^l_{i,\text{expert}}||_{2}+\epsilon))$
    \State $F^l_{i,\text{MoE}} \leftarrow \lambda_{MoE} \cdot F^l_{i,\text{norm}} + (1-\lambda_{MoE}) \cdot F^l_i$
\EndFor
\vspace{8pt}
%---------------------------------------------
\State $\mathcal{E}_{\text{tensor}} \leftarrow \text{Stack}(\mathcal{E})$ \Comment{Shape: $L \times K \times d$}
\State Compute ETF Loss $\mathcal{L}_{etf}$ from $\mathcal{E}_{\text{tensor}}$ (\cref{eq:etf_loss}).

\State \Return $F^l_{\text{MoE}},\ \mathcal{L}_{etf}$

\end{algorithmic}
\end{algorithm}

\section{Additional Information}
\label{sec:additional}
\subsection{Data Descriptions}
\label{sec:data}
We evaluated the zero-shot anomaly detection (ZSAD) performance of our proposed method on a total of 14 publicly available datasets, including five industrial and nine medical datasets, as presented in \cref{tab:dataset_overview}.
\begin{table*}[t!]
\centering
\caption{Summary of key statistics for the 14 publicly available industrial and medical datasets used in our study. It reports the number of classes, normal and anomalous images, and data types. The last column indicates the availability of evaluation (Pixel-level / Image-level), where 'X' denotes cases not applicable due to the absence of ground truth masks or normal images.}
\vspace{-2mm}
\footnotesize 

\begin{tabular*}{0.95\textwidth}{@{\extracolsep{\fill}}c|>{\centering\arraybackslash}p{1.8cm}|c c c c c@{}} 
\toprule

Domain & Dataset & \#Classes & \#Normal Images & \#Anomalous Images & Data Types & Evaluation (Pixel / Image) \\
\midrule
\multirow{5}{*}{\parbox{1.8cm}{\centering Industrial}} & MVTec & 15 & 467 & 1,258 & object \& texture & (O, O) \\
 & VisA & 12 & 962 & 1,200 & object & (O, O) \\
 & BTAD & 3 & 451 & 290 & object \& texture & (O, O) \\
 & RSDD & 1 & 387 & 387 & texture & (O, O) \\
 & DTD-synthetic & 12 & 375 & 947 & texture & (O, O) \\
\midrule
\multirow{9}{*}{\parbox{1.8cm}{\centering Medical}} & Brain MRI & 1 & 640 & 1,013 & brain & (O, O) \\
 & Head CT & 1 & 100 & 100 & brain & (X, O) \\
 & Liver CT & 1 & 833 & 660 & Liver & (O, O) \\
 & Retina OCT & 1 & 1,041 & 764 & Retina & (O, O) \\
 & ColonDB & 1 & 0 & 380 & Colon & (O, X) \\
 & ClinicDB & 1 & 0 & 612 & Colon & (O, X) \\
 & CVC-300 & 1 & 0 & 60 & Colon & (O, X) \\
 & Endo & 1 & 0 & 200 & Colon & (O, X) \\
 & Kvasir & 1 & 0 & 1,000 & Colon & (O, X) \\
\bottomrule
\end{tabular*}
\label{tab:dataset_overview} 
\end{table*}

\begin{itemize}
    \vspace{1ex}
	\item \textbf{MVTec-AD}~\cite{mvctec} is a benchmark dataset widely used for industrial anomaly detection. It comprises 15 object categories, with images captured by high-resolution industrial cameras. In this study, we use only the test set, which contains 467 normal and 1,258 anomalous images. Each anomalous image is accompanied by a pixel-level defect mask, enabling both image-level classification and pixel-level anomaly segmentation evaluation.

    \vspace{1ex}
    \item \textbf{VisA}~\cite{visa} is a benchmark dataset for industrial anomaly detection that includes surface and structural defects of manufactured products. It comprises 12 object categories, such as PCB, capsules, and candle. In this study, we employ only the test set, which contains 962 normal and 1,200 anomalous images. Each anomalous image provides a pixel-level defect mask, allowing both image-level classification and pixel-level anomaly segmentation evaluation.

    \vspace{1ex}
    \item \textbf{BTAD}~\cite{BTAD} is an industrial anomaly detection dataset collected from real manufacturing environments. It includes three product categories. In this study, we use only the test set, which contains 451 normal and 290 anomalous images.

    \vspace{1ex}
    \item \textbf{RSDD}~\cite{RSDD} is an industrial anomaly detection dataset designed for rail surface defect detection, consisting of two object-type categories. In this study, we use the test set, which includes 387 normal and 387 anomalous images. Pixel-level annotations are provided, allowing both image-level classification and pixel-level anomaly segmentation evaluation.

    \vspace{1ex}
    \item \textbf{DTD-Synthetic}~\cite{DTD-Synthetic} is a dataset generated by synthesizing data from 12 texture categories. Despite being synthetically created, it provides pixel-level anomaly masks, allowing evaluation at both the image and pixel levels. In this study, the dataset includes 375 normal images and 947 anomalous images.

    \vspace{1ex}
    \item \textbf{Brain MRI}~\cite{bmad} is a medical dataset consisting of brain MRI scans used for detecting various types of brain lesions. In this study, we use the test set, which includes 640 normal and 1,013 anomalous images. Each anomalous image is accompanied by a pixel-level defect mask, enabling both image-level classification and pixel-level anomaly segmentation evaluation.

    \vspace{1ex}
    \item \textbf{Liver CT}~\cite{bmad} is a medical dataset composed of liver CT scans. We utilize its test set, which consists of 833 normal and 660 anomalous images. Pixel-level segmentation masks are provided, enabling both image-level classification and pixel-level anomaly segmentation evaluation.

    \vspace{1ex}
    \item \textbf{Retina OCT}~\cite{bmad} is a medical dataset consisting of retinal images captured using Optical Coherence Tomography (OCT). We utilize its test set, which consists of 1,041 normal and 764 anomalous images. Pixel-level annotations of anomalous regions are provided, enabling both image-level and pixel-level evaluation.

    \vspace{1ex}
    \item \textbf{CVC-ColonDB}~\cite{ColonDB} is a medical anomaly detection dataset consisting of polyp images captured from colonoscopy procedures. The dataset contains 380 anomalous images, each accompanied by a pixel-level mask, but no normal samples. Due to the absence of normal data, we use this dataset exclusively for pixel-level anomaly segmentation evaluation.

    \vspace{1ex}
    \item \textbf{CVC-ClinicDB}~\cite{ClinicDB} is a medical anomaly detection dataset composed of polyp images obtained from colonoscopy examinations. Similar to CVC-ColonDB, it provides pixel-level segmentation masks and contains 612 anomalous images without any normal samples. Thus, it is used exclusively for pixel-level anomaly segmentation evaluation.

    \vspace{1ex}
    \item \textbf{CVC-300}~\cite{cvc300} is a medical anomaly detection dataset consisting of polyp images captured from colonoscopy procedures. It contains only 60 anomalous images with pixel-level annotations and no normal samples. Accordingly, we use this dataset exclusively for pixel-level anomaly segmentation evaluation.

    \vspace{1ex}
    \item \textbf{Endo}~\cite{Endo} is a medical anomaly detection dataset for polyp detection, comprising colon endoscopy images. It contains 200 anomalous images with pixel-level annotations and no normal samples. Consequently, this dataset is evaluated exclusively on pixel-level anomaly segmentation.

    \vspace{1ex}
    \item \textbf{Kvasir}~\cite{kvasir} is a medical anomaly detection dataset consisting of colon endoscopy images. It contains 1,000 anomalous images, each annotated with pixel-level masks for polyps, and includes no normal samples. In this study, we employ this dataset exclusively for pixel-level anomaly segmentation evaluation.

    \vspace{1ex}
    \item \textbf{Head CT}~\cite{HeadCT} is a medical dataset consisting of head CT scans used for detecting brain anomalies such as hemorrhages and tumors. In this study, we employ the refined version curated by AdaCLIP~\cite{cao2024adaclip}. The test set comprises 100 normal and 100 anomalous images. As this version provides only image-level labels, it is used exclusively for image-level anomaly classification. 
\end{itemize}

\subsection{Comparison Method Details}
\label{sec:method}
For a fair comparison, we unify the backbone to OpenCLIP ViT-L/14-336 across all baselines except WinCLIP~\cite{jeong2023winclip}, where we cite standard ViT-B/16$^{+}$-240 results~\cite{jeong2023winclip, qu2025bayesian} as the official code is unavailable. Furthermore, all competing models that require an auxiliary training phase are evaluated using weights trained on the VisA dataset under this unified setting (conversely, weights trained on MVTec-AD are used when evaluating on VisA).
\begin{itemize}
    \vspace{1ex}
	\item \textbf{WinCLIP}~\cite{jeong2023winclip} identifies that the original CLIP model is trained only on global embeddings, and its last visual feature maps before pooling are not well aligned with the language space, which limits its zero-shot segmentation performance. To address this issue, this paper proposes a window-based strategy. In this approach, the input image is divided into multi-scale sliding windows, and each masked image is independently passed through the CLIP image encoder to extract window-level global features that are aligned with the language space. In addition, a Compositional Prompt Ensemble is introduced, which combines state words and prompt templates to better define normal and anomalous conditions. We use the results of WinCLIP reported in~\cite{jeong2023winclip, qu2025bayesian}.

    \vspace{1ex}
	\item \textbf{APRIL-GAN}~\cite{chen2023april} points out that the patch-level image features in the original CLIP model are not properly mapped to a joint embedding space aligned with text features. To address this limitation, this paper introduces additional linear layers that map fine-grained patch features—extracted from multiple stages of the CLIP image encoder—into the joint embedding space. This approach enhances CLIP from a classification-only model to one capable of segmentation, allowing it to visually highlight where anomalies occur. We evaluate the model using the official pre-trained weights.

    \vspace{1ex}
    \item \textbf{AnomalyCLIP}~\cite{zhou2023anomalyclip} argues that the original CLIP model lacks the ability to distinguish between normal and abnormal concepts, which leads to degraded zero-shot anomaly detection (ZSAD) performance. To address this issue, this paper proposes an object-agnostic prompt learning strategy. Instead of explicitly including object class names, the method learns text prompts designed to capture generic indicators of normality and abnormality. These prompts are trained through a glocal context optimization process that combines global and local losses on an auxiliary dataset, while the visual encoder is fine-tuned using a Diagonally Prominent Attention Map (DPAM) to enhance local visual features. This approach allows the model to focus on shared anomaly patterns rather than object-specific semantics, enabling better generalization to unseen object categories. We evaluate the model using the official pre-trained weights.

    \vspace{1ex}
	\item \textbf{AdaCLIP}~\cite{cao2024adaclip} highlights the limitations of existing Static Prompt and Dynamic Prompt-based models, as well as the shortcomings of simple image-level detection approaches. To address these issues, this paper introduces a hybrid learnable prompts mechanism that combines static prompts—which provide basic adaptation for zero-shot anomaly detection (ZSAD)—with Dynamic Prompts that enable dynamic adaptation. These prompts are applied to both the image encoder and the text encoder. In addition, a Hybrid Semantic Fusion module is introduced to enhance image-level detection performance. We re-train the model using the official code and configuration.

    \vspace{1ex}
	\item \textbf{AA-CLIP}~\cite{ma2025aa} is the first to identify the unique anomaly unawareness problem of the original CLIP model — its inability to distinguish subtle semantic differences between normal and anomalous concepts. To address this issue, this paper proposes an effective two-stage adaptation strategy. In the first stage, the visual encoder is kept frozen, and the text encoder is adapted using a Residual Adapter and a Disentangle Loss to clearly separate normal and abnormal text anchors. In the second stage, the learned text anchors are fixed, and the visual encoder is fine-tuned with a Residual Adapter so that its patch-level features align with these text anchors. This controlled two-stage approach preserves CLIP’s strong generalization ability while efficiently injecting anomaly-aware information, thereby improving zero-shot performance. We re-train the model using the official code and configuration.

    \vspace{1ex}
	\item \textbf{Bayes-PFL}~\cite{qu2025bayesian} points out the excessive engineering required for manual prompt design and the limited generalization ability of simple learnable prompts. To address this issue, this paper proposes Bayesian Prompt Flow Learning, which models the text prompt space as a learnable probabilistic distribution from a Bayesian perspective. A Prompt Flow Module and a Residual Cross-Modal Attention Module are introduced to strengthen the alignment between dynamically generated text embeddings and fine-grained image patch features. We evaluate the model using the official pretrained weights.
\end{itemize}

\subsection{Detailed Experimental Setup}
\label{sec:setup}
We implement MoECLIP using the OpenCLIP ViT-L/14-336 architecture pre-trained by OpenAI~\cite{radford2021learning}, as our frozen backbone. All input images are uniformly resized to $518 \times 518$. We integrate our MoE modules at the output of the $l$-th layers of the Vision Encoder, where $l \in \{6, 12, 18, 24\}$, and text features are extracted from the final layer of the text encoder. During the training phase, we apply data augmentation~\cite{ma2025aa} with a 0.5 probability for each transformation, including random affine transformation, color jitter, random rotation, random horizontal flip, and random vertical flip. For the model configuration, we set the total number of experts $K$ to 4 with Top-2 routing via a linear router layer ${R}$, a LoRA rank $r$ to 8, a LoRA scaling factor $\alpha$ to 16, and a LoRA dropout rate of 0.05. We also set the MoE residual weight $\lambda_{MoE}$ to 0.1, and the normalization $\epsilon$ to $1 \times 10^{-6}$. The PAA module utilizes scales $s \in \{1, 3, 5\}$, and the auxiliary loss weights $\lambda_{etf}$ and $\lambda_{bal}$ are both set to 0.01. The projection layer used for aligning features to the text space is shared across all PAA scales within the same encoder layer. We train the model for 20 epochs using the Adam optimizer with $\beta_1$ set to 0.5, $\beta_2$ set to 0.999, and a batch size of 2. The initial learning rate is $5 \times 10^{-4}$, which is decayed by a factor of 0.5 at 16,000 and 32,000 iterations using a MultiStepLR scheduler. All experiments are conducted on 2 × NVIDIA Tesla V100 16GB GPUs and an Intel(R) Xeon(R) Gold 5120 CPU @ 2.20GHz, using PyTorch 2.3.1.

\subsection{Text Prompt}
\label{sec:prompt}
Following the methodology of AA-CLIP~\cite{ma2025aa}, we adopt identical prompt templates and descriptors as specified in \cref{tab:prompt_templates}. Throughout the training and inference phases, the "[\texttt{class}]" token is substituted with the class-specific description, while either the normal or anomaly descriptor is injected into the template.

\begin{table}[t!]
\centering
\footnotesize
\caption{Templates used for the Text Prompts}
\vspace{-2mm}
\renewcommand{\arraystretch}{1.0}
\begin{tabular}{l|l}
\hline
State & Prompt \\
\hline
Prompt Template & 
\begin{tabular}[c]{@{}l@{}}
\{\} \\
a photo of a \{\}
\end{tabular} \\
\hline
Normal Prompt &
\begin{tabular}[c]{@{}l@{}}
 {[\texttt{class}]} \\
the {[\texttt{class}]} \\
a {[\texttt{class}]}
\end{tabular} \\
\hline
Abnormal Prompt  &
\begin{tabular}[c]{@{}l@{}}
{[\texttt{class}]} with damage \\
{[\texttt{class}]} with defect \\
{[\texttt{class}]} with flaw \\
damaged {[\texttt{class}]} \\
broken {[\texttt{class}]} \\
\end{tabular} \\
\hline
\end{tabular}
\label{tab:prompt_templates}
\end{table}

\subsection{Details of Balance Loss}
\label{sec:balance}
In our MoE-based architecture, the router selects the optimal Top-k experts for each patch. A common failure mode is expert collapse, where the network favors a small subset of experts, which prevents genuine specialization because under-utilized experts receive few training signals. To mitigate this issue, we employ an auxiliary balance loss $\mathcal{L}_{bal}$~\cite{shazeer2017outrageously,meng2024moead}, based on the squared Coefficient of Variation ($\text{CV}^2$), that penalizes uneven load distribution. The $\text{CV}^2$ is computed from the batch load vector $B^l$, which represents the total routing weight assigned to each of the $K$ experts from all $L$ patches at layer $l$, as shown in \cref{eq:cv_calculation},
\begin{equation}
\label{eq:cv_calculation}
\text{CV}^2(B^l) = \frac{\sigma(B^l)^2}{\mu(B^l)^2 + \epsilon}, \quad \text{where } B^l = \sum_{i=1}^{L} R(F^l_i) 
\end{equation}
where $R(F^l_i) \in \mathbb{R}^K$ is the routing probability vector for the $i$-th patch; $\mu(B^l)$ and $\sigma(B^l)$ are the mean and standard deviation of the load vector $B^l$; and $\epsilon$ is a small constant ($1 \times 10^{-6}$) for numerical stability.
The final balance loss $\mathcal{L}_{bal}$ is the sum of this $\text{CV}^2$ value across all MoE layers, forcing the load distribution to be uniform by minimizing the variance, as shown in \cref{eq:total_balance_loss}:
\begin{equation}
\label{eq:total_balance_loss}
\mathcal{L}_{bal} = \sum_{l} \left[ \text{CV}^2\left(\sum_{i=1}^{L} R(F^l_i)\right) \right] = \sum_{l} \left[ \text{CV}^2(B^l) \right]
\end{equation}

{\small
\begin{figure*}[t]
    \centering
    \vspace{-2.5mm}
    \includegraphics[width=1\linewidth]{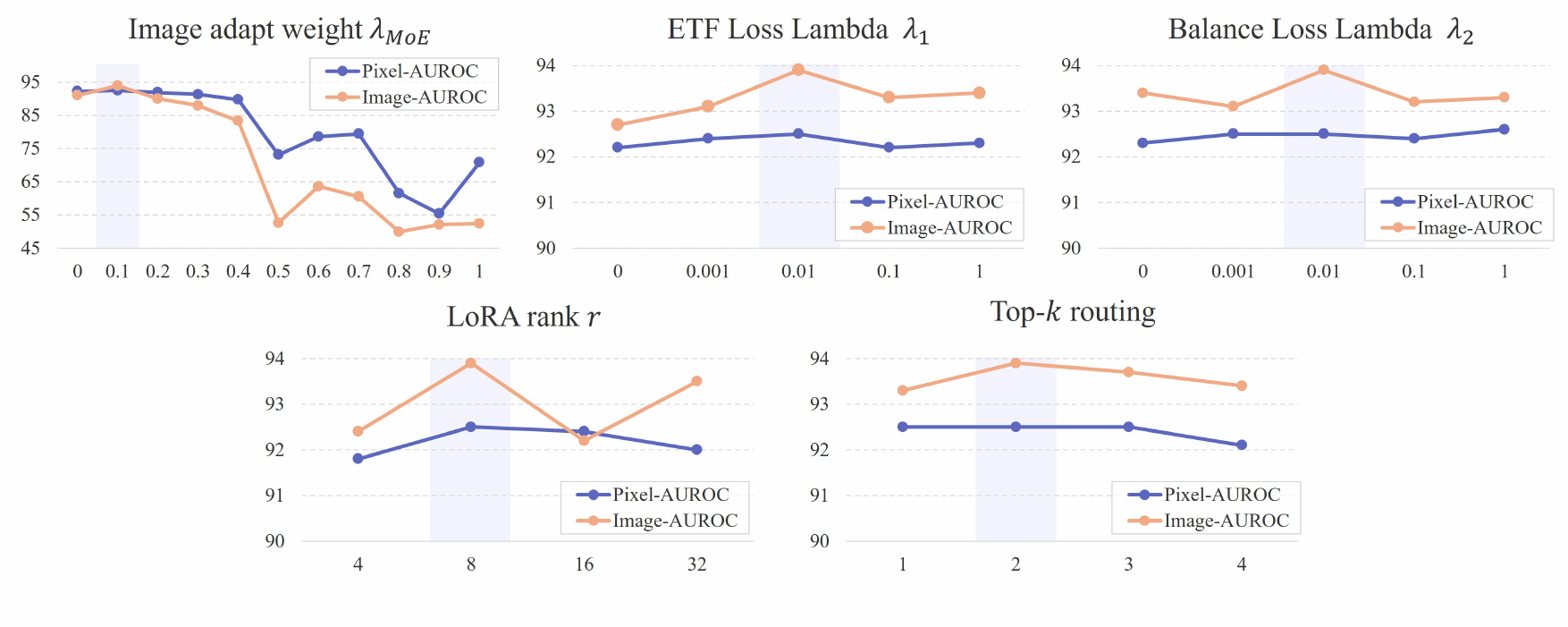}
    \vspace{-10mm}
    \caption{Ablation study of hyperparameters on the MVTec-AD dataset.}
    \label{fig:ablation_hyperparameter}
    \vspace{-2.5mm}
\end{figure*}
}

\subsection{Details of Focal Loss and Dice Loss}
\label{sec:loss_detail}

\vspace{1ex}
\noindent\textbf{Focal Loss: }
In anomaly segmentation, the extreme imbalance between vast normal regions and small anomalous areas causes standard Cross-Entropy loss to be overwhelmed by easy negative samples. This issue is mitigated by the Focal Loss~\cite{focalloss}, which down-weights the contribution of these high-confidence predictions to effectively focus learning on misclassified or ambiguous samples. The Focal loss is defined as:
\begin{equation}
\label{eq:focal_loss} 
\mathcal{L}_{Focal} = -\frac{1}{N} \sum_{i=1}^{N} (1 - p_i)^\gamma \log(p_i) 
\end{equation}
where $N$ is the total number of pixels, $p_i$ is the predicted probability, and $\gamma$ is the focusing parameter that controls the down-weighting rate of easy negative samples. As $\gamma$ increases, the loss contribution of well-classified pixels ($p_i \to 1$) is suppressed more aggressively, forcing the model to focus on hard, misclassified pixels with low predicted probabilities ($p_i \to 0$). In our implementation, we set $\gamma$ to 2.

\vspace{1ex}
\noindent\textbf{Dice Loss: }
Complementing Focal Loss, the Dice Loss~\cite{diceloss} is employed to directly optimize the overlap between the predicted anomaly map and the ground-truth mask. It is particularly effective for small anomalous regions, as it is insensitive to the dominance of large background regions. The Dice loss is defined as:
\begin{equation} 
\label{eq:dice_loss}
\mathcal{L}_{Dice} = 1 - \frac{2 \sum_{i=1}^{N} y_i \hat{y}_i}{\sum_{i=1}^{N} y_i + \sum_{i=1}^{N} \hat{y}_i} 
\end{equation}
where $y_i$ and $\hat{y}_i$ denote the ground truth and the predicted probability for the $i$-th pixel in the image, respectively. Minimizing this loss effectively maximizes the overlap between the prediction and the ground truth.

\section{Additional Experimental Analysis}
\label{sec:experiment}

\subsection{Analysis of Model Configurations}
\label{sec:more_configuration}

\vspace{1ex}
\noindent\textbf{Analysis of Model Hyperparameters: }
\cref{fig:ablation_hyperparameter} presents the results of our model hyper-parameter analysis on the MVTec-AD dataset. We first analyze the image adapt weight $\lambda_{\text{MoE}}$, which balances the original CLIP feature $F^l_i$ against the adapted MoE feature $F^l_{i,\text{norm}}$ (\cref{eq:final_moe_output}). A small value of 0.1 yields the best performance, while relying more heavily on the adapted features via larger $\lambda_{\text{MoE}}$ values causes a significant performance degradation after 0.4. This strongly suggests that relying too heavily on the adapted features damages the vital generalization capability preserved in the original frozen CLIP features.

For the auxiliary loss weights, both $\lambda_{1}$ (ETF) and $\lambda_{2}$ (Balance) achieve optimal results at 0.01. Disabling these losses or weighting them too heavily degrades ZSAD performance, confirming that a small amount of regularization is crucial for expert specialization and stability.

The LoRA rank $r$ analysis shows that performance peaks at $r=8$. Increasing the rank does not guarantee better performance. This indicates that $r=8$ provides sufficient capacity, while a higher rank may increase the risk of overfitting to the auxiliary dataset.

The Top-k analysis using a total of 4 experts demonstrates that $k=2$ (Top-2 routing) clearly outperforms other values. This finding suggests that using more experts can be detrimental to ZSAD performance, possibly by introducing noise or redundant information, whereas $k=2$ provides the optimal balance of specialized features.
\par
\begin{table}[t]
\centering
\caption{Ablation study on different backbones, image resolutions, and patch sizes, evaluated on the MVTec-AD dataset. The best performance is in \textbf{bold}.}
\vspace{-2mm}
\resizebox{0.475\textwidth}{!}{
\begin{tabular}{cccccccc}
\toprule
\multirow{2}{*}{Backbone} & \multirow{2}{*}{Patch size} & \multirow{2}{*}{Image Resolution} &
& \multicolumn{2}{c}{Image-level} & \multicolumn{2}{c}{Pixel-level} \\ 
\cmidrule(lr){5-6} \cmidrule(lr){7-8}
& & & & AUROC & AP & AUROC & AP \\ 
\midrule
ViT-B/32-224 & 32 & $448 {\times} 448$ & & 82.7 & 91.8 & 88.4 & 28.5 \\
ViT-B/16-224 & 16 & $448 {\times} 448$ & & 87.9 & 94.7 & 90.9 & 39.3 \\
ViT-L/14-224 & 14 & $518 {\times} 518$ & & 91.8 & 95.8 & 91.3 & 42.7 \\
ViT-L/14-336 & 14 & $448 {\times} 448$ & & 93.1 & 96.7 & 91.0 & 45.2 \\
\textbf{ViT-L/14-336} & \textbf{14} & \boldmath{$518 \times 518$} & & \textbf{93.9} & \textbf{96.8} & \textbf{92.5} & \textbf{45.7} \\
ViT-L-14-336 & 14 & $602 {\times} 602$ & & 93.3 & 96.7 & 91.1 & 45.4 \\
\bottomrule
\end{tabular}
}
\label{tab:backbone}
\vspace{-2.5mm}
\end{table}

\vspace{1ex}
\noindent\textbf{Analysis of Backbone, Resolution, and Patch Size: }
In \cref{tab:backbone}, we analyze the influence of backbone, patch size, and image resolution. First, a clear trend emerges with patch size: decreasing the patch size from 32 (ViT-B/32) to 16 (ViT-B/16) significantly improves performance. This suggests that smaller patches provide more detailed local information, which is crucial for identifying anomalous regions.
Similarly, scaling the model backbone from ViT-B to ViT-L also yields a substantial performance increase, confirming the benefit of a larger model capacity. Regarding image resolution, performance generally improves when increasing the size from $448 \times 448$ to $518 \times 518$ using the ViT-L/14-336 model. However, we observe a performance drop when increasing the resolution further to $602 \times 602$. This indicates a potential trade-off, suggesting that resolutions significantly larger than the model's pre-training resolution ($336\times336$) may not be optimal. Based on this analysis, the ViT-L/14-336 model using a $518 \times 518$ input resolution achieves the best performance. Therefore, we select this configuration as the default backbone for MoECLIP.

\vspace{1ex}
\noindent\textbf{Fixed-Prompt Ensemble vs. Learnable Prompt: }In \cref{tab:learnable_prompt}, we investigate the impact of our text prompt design. The default MoECLIP configuration utilizes an ensembling strategy based on a set of fixed base prompts (e.g., ``a photo of a'') combined with state-specific words (``damaged'', ``broken'', etc.) and the class name.
We compare this fixed prompt ensemble strategy against an alternative design based on prompt tuning, where the base prompts are replaced with learnable parameters.
The results indicate that our fixed-prompt design achieves higher AUROC/AP scores than the learnable-parameter variant at both the image and pixel levels. This suggests that while prompt tuning offers flexibility, it may be prone to overfitting on the auxiliary dataset. Our use of a fixed-prompt ensemble appears to better preserve the generalization capabilities of the original CLIP model, leading to superior ZSAD performance.

\begin{table}[H]
\centering
\caption{Comparison of fixed vs. learnable text prompt strategies. The table compares our default fixed-prompt ensemble against a learnable prompt tuning variant, evaluated on the MVTec-AD dataset. The best performance is in \textbf{bold}.}
\vspace{-2mm}
\resizebox{0.475\textwidth}{!}{
\begin{tabular}{cccccc}
\toprule
\multirow{2}{*}{Method} & \multirow{2}{*}{Prompt} &
\multicolumn{2}{c}{Image-level} & \multicolumn{2}{c}{Pixel-level} \\ 
\cmidrule(lr){3-4} \cmidrule(lr){5-6}
& & AUROC & AP & AUROC & AP \\ 
\midrule
MoECLIP (Fixed) & 
\begin{tabular}[c]{@{}c@{}}
$t_n = \text{a photo of a } [\texttt{class}]$ \\
$t_a = \text{a photo of a } [\texttt{state}] [\texttt{class}]$
\end{tabular} 
& \textbf{93.9} & \textbf{96.8} & \textbf{92.5} & \textbf{45.7} \\
MoECLIP (Learnable) & 
\begin{tabular}[c]{@{}c@{}}
$t_n = [V_1][V_2][V_3][V_4][\texttt{class}]$ \\
$t_a = [W_1][W_2][W_3][W_4][\texttt{state}][\texttt{class}]$
\end{tabular} 
& 92.2 & 96.3 & 91.9 & 44.1 \\
\bottomrule
\end{tabular}
}
\label{tab:learnable_prompt}
\vspace{-2.5mm}
\end{table}

\vspace{1ex}
\noindent\textbf{Analysis of MoE Module Placement: }In \cref{tab:moe_layer_selection}, we analyze the effect of placing MoE modules at different ViT layers. We observe that integrating modules in a concentrated block—either at the shallow layers (e.g., 1–4) or the deep layers (e.g., 20–24)—yields suboptimal results. Similarly, a dense integration across 8 layers (e.g., 3, 6, 9...24) also underperforms, particularly in Pixel-Level AUROC.
In contrast, our chosen strategy of sparsely integrating modules at four key layers (6, 12, 18, and 24) achieves superior AUROC and AP scores at both the image and pixel levels. These results suggest that integrating MoE capacity strategically across a few key stages leverages complementary representations from different depths more effectively than either concentrated or overly dense approaches.

\begin{table}[htbp]
\centering
\caption{Ablation study on the layer placement of MoE modules. The best performance is in \textbf{bold}.}
\vspace{-2mm}
\renewcommand{\arraystretch}{0.85}
\footnotesize 
\begin{tabular*}{\columnwidth}{c @{\extracolsep{\fill}} cc cc} 
\toprule
\multirow{2}{*}[-2pt]{MoE Integration Layers} & \multicolumn{2}{c}{Image-Level} & \multicolumn{2}{c}{Pixel-Level} \\
\cmidrule(lr){2-3} \cmidrule(lr){4-5}
& AUROC & AP & AUROC & AP \\
\midrule
1, 2, 3, 4 & 91.0 & 96.2 & 92.2 & 43.6 \\
20, 21, 22, 23, 24 & 91.8 & 96.6 & 91.6 & 43.4 \\
3, 6, 9, 12, 15, 18, 21, 24 & 92.3 & 96.5 & 90.9 & 45.3 \\
\midrule
\textbf{6, 12, 18, 24} & \textbf{93.9} & \textbf{96.8} & \textbf{92.5} & \textbf{45.7} \\
\bottomrule
\end{tabular*}
\label{tab:moe_layer_selection}
\vspace{-2.5mm}
\end{table}

\vspace{1ex}
\noindent\textbf{Single-layer vs. Multi-layer Features: }In \cref{tab:patch_feature}, we further examine the contribution of features extracted from individual ViT layers compared to a multi-layer ensemble. Using a single layer yields highly different performance characteristics depending on depth: early layers (e.g., Layer 6) provide limited semantic discrimination, resulting in lower image-level AUROC and AP scores, while deeper layers (e.g., Layers 12, 18, and 24) yield progressively stronger results due to richer semantic abstractions. However, none of the single-layer configurations surpass the multi-layer ensemble, which aggregates features from Layers 6, 12, 18, and 24. The ensemble achieves the best performance across both image-level and pixel-level metrics, demonstrating that complementary information from multiple hierarchical stages is essential for robust anomaly detection. These results highlight the benefit of leveraging multi-layer ViT features rather than relying on a single feature depth.

\begin{table}[ht!]
\centering
\caption{Ablation study on the multi-layer feature ensemble. The table compares our 'Ensemble' method (aggregating 4 layers) against configurations using features from only a single ViT layer (6, 12, 18, or 24). The best performance is in \textbf{bold}.}
\vspace{-2mm}
\renewcommand{\arraystretch}{0.85}
\footnotesize
\begin{tabular*}{\linewidth}{@{\extracolsep{\fill}}c cc cc@{}}
\toprule
\multirow{2}{*}[-1pt]{Layers} 
& \multicolumn{2}{c}{Image-Level} 
& \multicolumn{2}{c}{Pixel-Level} \\
\cmidrule(lr){2-3} \cmidrule(lr){4-5}
& AUROC & AP & AUROC & \multicolumn{1}{c}{AP} \\ 
\midrule
\multicolumn{1}{c}{Layer 6}  & 76.0 & 87.4 & 83.4 & \multicolumn{1}{c}{28.9} \\
\multicolumn{1}{c}{Layer 12} & 90.0 & 95.0 & 91.7 & \multicolumn{1}{c}{43.7} \\
\multicolumn{1}{c}{Layer 18} & 92.0 & 96.3 & 91.0 & \multicolumn{1}{c}{43.4} \\
\multicolumn{1}{c}{Layer 24} & 91.6 & 95.7 & 91.4 & \multicolumn{1}{c}{42.6} \\
\midrule
\multicolumn{1}{c}{\textbf{Ensemble}}
& \textbf{93.9} & \textbf{96.8} & \textbf{92.5} & \multicolumn{1}{c}{\textbf{45.7}} \\
\bottomrule
\end{tabular*}
\label{tab:patch_feature}
\end{table}

\vspace{1ex}
\noindent\textbf{Analysis of PAA Scale Configuration: }
In \cref{tab:paa_selection}, we analyze the impact of scale configurations in Patch Average Aggregation (PAA). The original feature ($s=1$) provides the baseline performance, whereas using single larger scales ($s=3$ or $5$) degrades performance by obscuring fine-grained details. Conversely, the multi-scale combination ($s \in \{1, 3, 5\}$) achieves the highest accuracy, confirming that integrating local details ($s=1$) with contextual fields ($s=3, 5$) is essential. Thus, a balanced combination of the original feature and mid-range scales is optimal.

\begin{table}[htbp]
\centering
\caption{Ablation study on the multi-scale configuration of PAA. The best performance is in \textbf{bold}.}
\vspace{-2mm}
\renewcommand{\arraystretch}{0.85} 
\footnotesize
\begin{tabular*}{\columnwidth}{c @{\extracolsep{\fill}} cc cc}
\toprule
\multirow{2}{*}[-2pt]{PAA Scale $s$} 
& \multicolumn{2}{c}{Image-Level} 
& \multicolumn{2}{c}{Pixel-Level} \\
\cmidrule(lr){2-3} \cmidrule(lr){4-5}
& AUROC & AP & AUROC & AP \\
\midrule
1           & 92.8 & 96.0 & 92.1 & 44.1 \\
3           & 92.3 & 96.5 & 91.2 & 42.8 \\
5           & 91.8 & 96.5 & 91.5 & 43.2 \\
1, 3        & 90.7 & 95.2 & 92.4 & 45.2 \\
1, 5        & 92.2 & 96.5 & 91.9 & 45.2 \\
3, 5        & 92.0 & 96.4 & 90.0 & 43.7 \\
\midrule
\textbf{1, 3, 5} & \textbf{93.9} & \textbf{96.8} & \textbf{92.5} & \textbf{45.7} \\
\bottomrule
\end{tabular*}
\label{tab:paa_selection}
\end{table}

\vspace{1ex}
\noindent\textbf{Analysis of MoE Configurations: }
We analyze robustness across router design, expert capacity, and expert constraint mechanisms in \cref{tab:configuration}. Increasing capacity via deeper MLP routers or full-rank experts leads to overall performance degradation, suggesting reduced generalization. Alternative constraints on experts (contrastive~\cite{feng2025comoe}, cosine similarity~\cite{cui2025cmoa}) also underperform our FOFS+ETF. In \cref{tab:perturbation}, we assess router stability by injecting adaptive Gaussian noise $\epsilon \sim \mathcal{N}(0, (\alpha \cdot \sigma)^2)$ into routing logits, where $\sigma$ is the logit standard deviation. Even at $\alpha=1.0$, the performance remains stable with Pixel-level AUROC exhibiting zero degradation (maintaining 92.5) and Image-level AUROC dropping by only a marginal 0.5. This strong resilience indicates that the router forms highly confident and sharp routing distributions, where expert assignment decisions are not easily flipped by logit noise. These results confirm MoECLIP's robustness to diverse configurations and internal perturbations.

\begin{table}[htbp]
\centering
\caption{Robustness analysis of expert configurations and router designs on the MVTec-AD dataset evaluated by (AUROC, AP). The best performance is in \textbf{bold}.}
\vspace{-2mm}
\label{tab:configuration}
\renewcommand{\arraystretch}{0.85}
\footnotesize
\begin{tabular*}{\columnwidth}{c @{\extracolsep{\fill}} cc cc}
\toprule
\multirow{2}{*}{Method}
& \multicolumn{2}{c}{Image-Level}
& \multicolumn{2}{c}{Pixel-Level} \\
\cmidrule(lr){2-3} \cmidrule(lr){4-5}
& AUROC & AP & AUROC & AP \\
\midrule
MLP Router
& 93.3 & \textbf{97.1}
& 92.1 & 45.5 \\
Full-Rank Expert
& 92.8 & 96.7
& 91.9 & \textbf{46.0} \\
Contrastive Constraint
& 92.4 & 96.2
& 91.7 & 45.6 \\
Similarity Constraint
& 92.7 & 96.7
& 91.9 & 45.5 \\
\midrule
\textbf{MoECLIP}
& \textbf{93.9} & 96.8
& \textbf{92.5} & 45.7 \\
\bottomrule
\end{tabular*}
\end{table}

\begin{table}[htbp]
\centering
\caption{Robustness of the routing mechanism to logit perturbation on the MVTec-AD dataset evaluated by AUROC and AP.}
\vspace{-2mm}
\label{tab:perturbation}
\renewcommand{\arraystretch}{0.85}
\footnotesize
\begin{tabular*}{\columnwidth}{c @{\extracolsep{\fill}} cc cc}
\toprule
\multirow{2}{*}{Perturbation ($\alpha$)} & \multicolumn{2}{c}{Image-Level} & \multicolumn{2}{c}{Pixel-Level} \\
\cmidrule(lr){2-3} \cmidrule(lr){4-5}
& AUROC & AP & AUROC & AP \\
\midrule
0.0 & 93.8 & 96.8 & 92.5 & 45.7 \\
0.2 & 93.6 & 96.7 & 92.5 & 45.7 \\
0.4 & 93.7 & 96.7 & 92.5 & 45.6 \\
0.6 & 93.7 & 96.7 & 92.5 & 45.5 \\
0.8 & 93.5 & 96.7 & 92.5 & 45.6 \\
1.0 & 93.3 & 96.4 & 92.5 & 45.6 \\
\bottomrule
\end{tabular*}
\end{table}

\subsection{Analysis of Training on Medical Data}
\label{sec:aux_medical}
While MoECLIP trained on the industrial dataset demonstrates strong generalization performance to unseen medical datasets, a performance gap due to the industrial-medical domain shift remains. This suggests that ZSAD performance can be further enhanced by leveraging an auxiliary dataset from the same medical domain. To investigate this, we train an additional MoECLIP model using the Brain MRI~\cite{HeadCT} dataset as the auxiliary source, as it provides both category labels and segmentation ground truths. The results are presented in \cref{tab:medical_ablation_v2}, where the model is evaluated on the 8 other medical datasets excluding Brain MRI itself. On average, while the medical-domain model shows a modest improvement in mean AUROC ($1.8\%$ for Image-level and $0.6\%$ for Pixel-level), the most significant gains are in Average Precision (AP). The model improves the mean Image-level AP by $3.4\%$ and the Pixel-level AP by a substantial $6.7\%$ across the 8 medical datasets. This confirms that using an auxiliary dataset from the same domain is an effective strategy for enhancing ZSAD performance. The large gains in AP, which is particularly sensitive to the precision of localization and false positives, suggest that in-domain training primarily improves the model's confidence and accuracy in identifying true anomalous regions.

\begin{table}[htbp]
\centering
\caption{Comparison of MoECLIP's ZSAD performance on 8 unseen medical datasets, when trained on a same-domain medical auxiliary dataset (Brain MRI) versus an out-of-domain industrial dataset (VisA). The best performance is in \textbf{bold}.}
\vspace{-2mm}
\resizebox{0.475\textwidth}{!}{
\begin{tabular}{c cc cc}
\toprule
& \multicolumn{2}{c}{MoECLIP (Industrial)} & \multicolumn{2}{c}{MoECLIP (Medical)} \\
\cmidrule(lr){2-3} \cmidrule(lr){4-5}

\multirow{2}{*}[7pt]{Dataset} & Image-level & Pixel-level & Image-level & Pixel-level \\
& (AUROC, AP) & (AUROC, AP) & (AUROC, AP) & (AUROC, AP) \\
\midrule
Head CT & (96.6, 94.5) & -- & (98.8, 99.0) & -- \\
Liver CT & (74.0, 64.6) & (97.2, 10.8) & (77.2, 70.8) & (97.7, 10.3) \\
Retina OCT & (85.5, 84.9) & (96.2, 66.3) & (85.7. 84.4) & (95.9, 60.1) \\
ColonDB & -- & (85.4, 34.8) & -- & (88.7, 52.1) \\
ClinicDB & -- & (89.7, 49.9) & -- & (90.4, 57.9) \\
CVC-300 & -- & (97.0, 53.0) & -- & (97.8, 73.0) \\
Endo & -- & (91.0, 62.5) & -- & (90.3, 65.3) \\
Kvasir & -- & (88.1, 57.6) & -- & (88.4, 62.7) \\
\midrule
Average & (85.4, 81.3) & (92.1, 47.8) & \textbf{(87.2, 84.7)} & \textbf{(92.7, 54.5)} \\
\bottomrule
\end{tabular}
}
\label{tab:medical_ablation_v2}
\end{table}

\subsection{Analysis of Computation Overhead}
\label{sec:computation}

As analyzed in \cref{tab:computation_overhead}, patch-level routing entails a modest overhead (+39ms, 0.2G FLOPs). However, our lightweight LoRA experts and sparse Top-k routing strategy reduce peak memory by 34.3\% and parameters by 1.7\% compared to AA-CLIP (single expert)~\cite{ma2025aa}, while boosting AUROC on MVTec-AD by 3.0\% and 0.9\% at the image and pixel levels, respectively, offering a favorable trade-off for ZSAD.

\begin{table}[htbp]
\centering
\caption{Evaluation of computation overhead and ZSAD performance (AUROC, AP).}
\vspace{-2mm}
\label{tab:computation_overhead}
\resizebox{0.98\columnwidth}{!}{
\renewcommand{\arraystretch}{0.95}
\begin{tabular}{c c c c c c c}
\toprule
\multirow{2}{*}{\makecell{Model}}
& \multirow{2}{*}{\makecell{Params (M)}}
& \multirow{2}{*}{\makecell{Inference\\ Time (ms)}}
& \multirow{2}{*}{\makecell{Peak GPU\\ Memory (MB)}}
& \multirow{2}{*}{\makecell{FLOPs (G)}}
& \multicolumn{2}{c}{\makecell{MVTec-AD}} \\
\cmidrule(lr){6-7}
& & & & & Image-level & Pixel-level \\
\midrule
\makecell{AA-CLIP\\(Single Expert)}
& 441.3
& \textbf{119.6}
& 13,853.8
& \textbf{1120.6}
& (90.9, 96.0)
& (91.6, 45.4) \\
\midrule
\textbf{MoECLIP}
& \textbf{433.6}
& 158.9
& \textbf{9,096.4}
& 1120.8
& (\textbf{93.9}, \textbf{96.8})
& (\textbf{92.5}, \textbf{45.7}) \\
\bottomrule
\end{tabular}
}
\end{table}

\section{Theoretical Grounding for MoE in ZSAD}
\label{sec:Theoretical}

Zero-Shot Anomaly Detection (ZSAD) requires capturing highly diverse
and heterogeneous visual primitives from auxiliary data to generalize
to unseen anomalies.
However, when adapting a monolithic model to learn these diverse
patterns simultaneously, the network frequently encounters
optimization challenges due to gradient conflicts.
Let $\theta$ denote the shared parameters of a monolithic network,
and let $\ell_{i}$ and $\ell_{j}$ be the patch-specific losses for
distinct patch features $F_{i}^{l}$ and $F_{j}^{l}$.
The gradient conflict can be mathematically expressed as:
\begin{equation}
    \cos\!\left(\nabla_{\theta}\ell_{i},\,\nabla_{\theta}\ell_{j}\right) < 0.
\end{equation}
This negative cosine similarity indicates destructive interference
during backpropagation, which impedes model convergence and degrades
generalization performance.

To fundamentally address this optimization bottleneck, MoECLIP
employs a Mixture-of-Experts (MoE) architecture that decouples the
parameter space.
For a given LoRA-based expert $n$ with a frozen down-projection
matrix $A_{n}$ and a learnable up-projection matrix $B_{n}$, the
gradient of the total loss $\mathcal{L}$ with respect to $B_{n}$ is
formulated through conditional routing:
\begin{equation}
    \nabla_{B_{n}}\mathcal{L}
    = \sum_{i} \hat{R}_{n}(F_{i}^{l})\,
      \frac{\partial\ell_{i}}{\partial\mathbf{y}_{i,n}}\,
      (A_{n}F_{i}^{l})^{\top},
\end{equation}
where $\hat{R}_{n}(F_{i}^{l})$ is the routing probability assigned to
expert $n$ for patch $F_{i}^{l}$, and
$\mathbf{y}_{i,n} = B_{n}A_{n}(F_{i}^{l})$ is the output of expert $n$
for patch $i$.
This formulation demonstrates that the parameter update for each
expert is strictly conditioned on its routing weights, effectively
isolating the gradients of heterogeneous patches.

FOFS further strengthens this isolation by
assigning each expert $n$ to a disjoint input subspace $c_n$.
Because $A_n$ extracts only the $c_n$-partition of the input, the
effective gradient signal $(A_{n}F_{i}^{l})$ for expert $n$ is
computed from a strictly non-overlapping region of the feature
dimension, independent of any other expert $m \neq n$.
This structural information separation at the input level ensures
that gradient updates across experts do not interfere, regardless of
their inner-product relationship. Crucially, FOFS enforces orthogonality only on LoRA A matrices (input projections), not on outputs. After orthogonal projection, each expert applies distinct learnable B matrices and outputs are aggregated via gating This architectural design enables the final representation to freely model the non-orthogonal, overlapping feature interactions necessary for detecting multi-faceted anomalies. Orthogonality in the input subspace serves as a beneficial inductive bias for specialization without limiting the model's representational flexibility.
Complementarily, the ETF constraint prevents experts from collapsing
into redundant representations in the output space by maximizing the
pairwise angle between the expert outputs $\hat{e}_{i,n}$ and $\hat{e}_{i,m}$. 

This synergistic design---FOFS for input-level information separation
and ETF for output-level diversity---mitigates gradient interference
at both stages. ~\cref{fig:convergence} empirically validates that MoECLIP converges 
1.3$\times$ faster than the Single Expert model, demonstrating that 
mitigating destructive gradient interference leads to more stable and 
efficient optimization in the ZSAD setting.

{\small
\begin{figure}[t]
    \centering
    \vspace{-2.5mm}
    \includegraphics[width=1.0\linewidth]{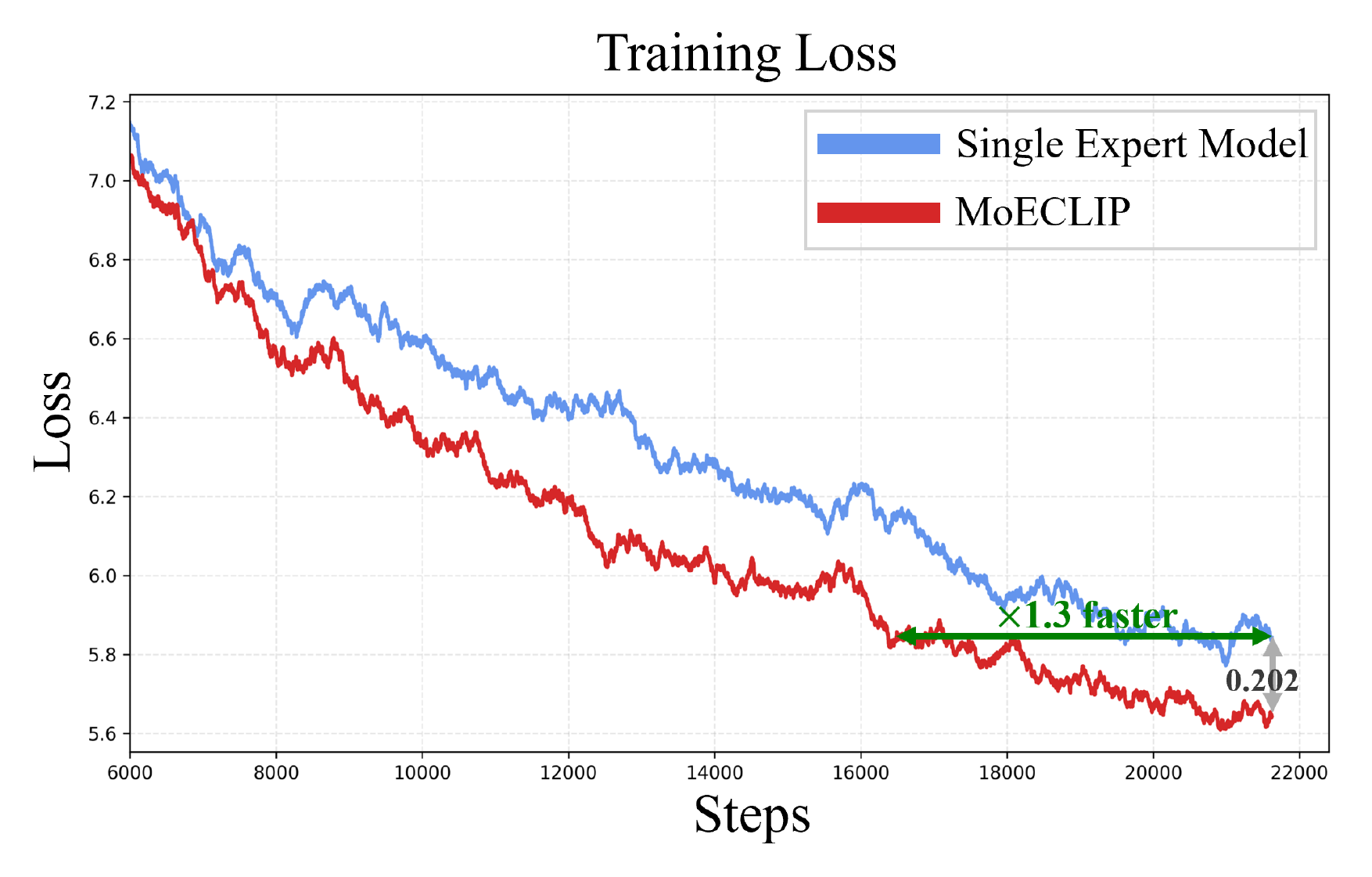}
    \vspace{-5mm}
    \caption{Comparison of training loss curves between the Single Expert Model and MoECLIP.}
    \label{fig:convergence}
\end{figure}
}

\section{Analysis of Expert Specialization}
\subsection{Expert Utilization}
\label{sec:Utilization}

The low activation of Expert 4 in \cref{fig:gradcam} reflects dataset-specific characteristics rather than redundancy. MVTec-AD inherently lacks the specific patterns that Expert 4 specializes in. \cref{fig:Utilization_graph} shows that expert utilization varies across datasets, demonstrating that each expert captures distinct patch characteristics. This confirms specialization rather than over-parameterization. Since ZSAD targets unseen datasets with unpredictable characteristics, retaining experts that may be underutilized on specific datasets is crucial for robust generalization.

{\small
\begin{figure}[ht]
    \centering
    \vspace{-2.5mm}
    \includegraphics[width=1\linewidth]{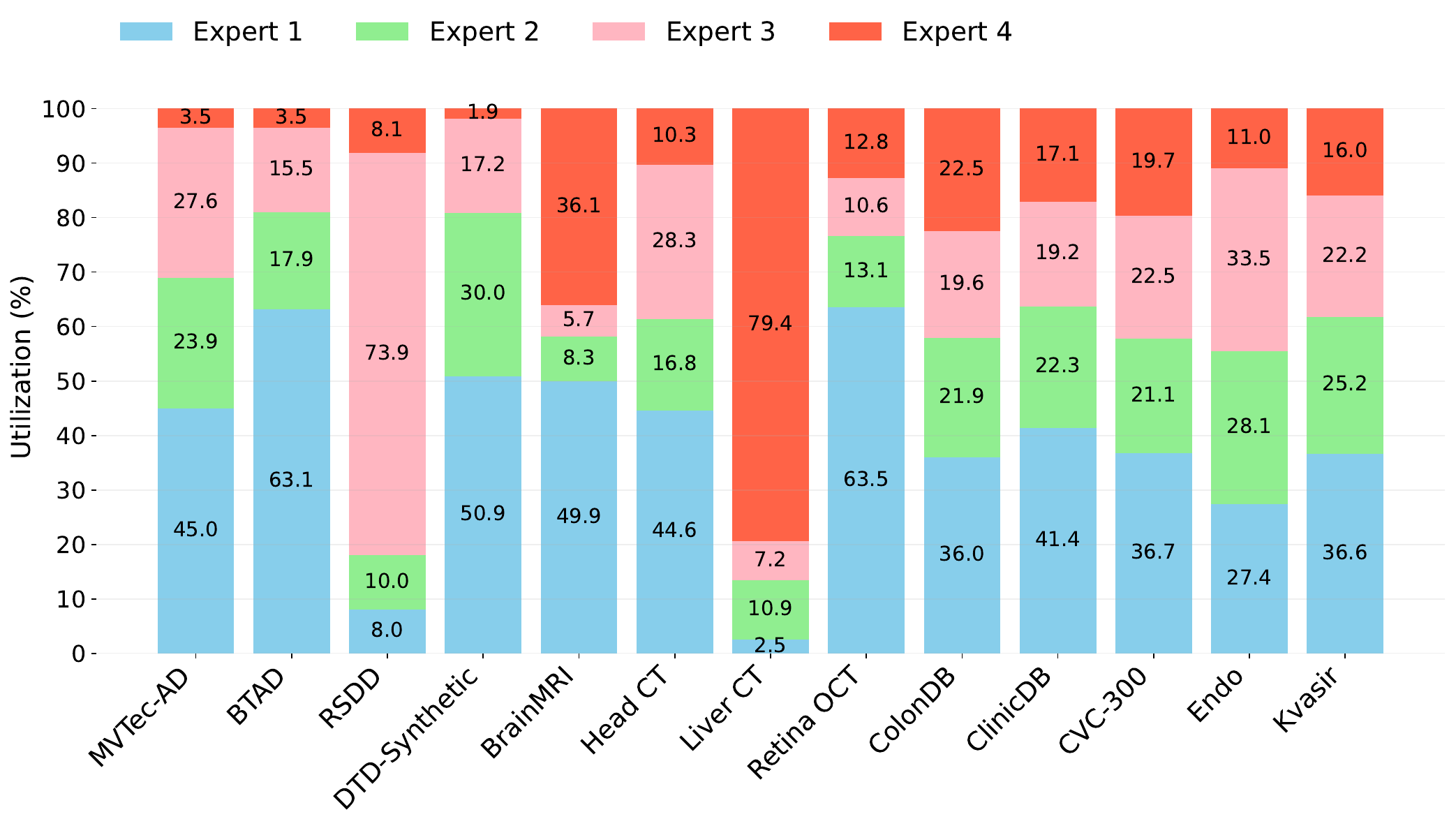}
    \vspace{-5mm}
    \caption{Distribution of Top-2 expert utilization (\%) at layer 18. The stacked bars illustrate the average proportion of routing assignments among the four experts for each dataset.}
    \label{fig:Utilization_graph}
\end{figure}
}

\subsection{Expert Interpretability}
\label{sec:Interpretability}

We quantitatively validated expert specialization by measuring the average visual properties of patches assigned to each expert: Sobel Gradient (edge strength), Contrast (pixel std), and Shannon Entropy (pattern complexity).
\cref{tab:interpretability} confirms the experts on BrainMRI exhibit functional differentiation. Expert 4 captures the non-informative background (near-zero stats), while Expert 1 attends to homogeneous regions with low variability. In contrast, Expert 2 specializes in strong structural edges (Highest Gradient) and Expert 3 focuses on high-complexity patterns (Highest Entropy). This confirms that MoECLIP disentangles patches into distinct visual features—ranging from background to structural edges and complex patterns—providing the quantitative explainability crucial for medical domains.

\begin{table}[ht]
\centering
\caption{Quantitative analysis of expert specialization using 
low-level visual features at layer 18 on BrainMRI (Top-1 selection).}
\vspace{-2mm}
\label{tab:interpretability}
\resizebox{1.0\columnwidth}{!}{
\renewcommand{\arraystretch}{1.2}
\begin{tabular}{l c c c l}
\toprule
\makecell{Expert} & \makecell{Gradient ($\nabla$)} & \makecell{Contrast ($\sigma$)} & \makecell{Entropy ($H$)} & \makecell{Specialization} \\
\midrule
\makecell{Expert 1} & 48.7 & 17.7 & 1.7 & \makecell{Low-variance regions} \\
\makecell{Expert 2} & 152.2 & 66.9 & 5.4 & \makecell{Structural edges} \\
\makecell{Expert 3} & 137.0 & 60.9 & 5.7 & \makecell{Complex patterns} \\
\makecell{Expert 4} & 0.7 & 0.2 & 0.03 & \makecell{Background} \\
\bottomrule
\end{tabular}
}
\vspace{-0.6em}
\end{table}%\vspace{0.5ex}

\section{Detailed Quantitative Results}
\label{sec:quantitative}

\subsection{Performance Results for All Categories}
\label{sec:all_categories}
In \cref{tab:full_results}, we report the detailed class-wise performance on the MVTec, VisA, BTAD, and DTD-Synthetic datasets.

\subsection{Inter-Expert Similarity}
\label{sec:similarity_expert}
\cref{fig:all_heatmaps} provides a comparison of inter-expert cosine similarity for both the Original MoE baseline lacking FOFS \&\ ETF Loss and our MoECLIP model across layers 6, 12, 18, and 24. This matrix is computed by averaging the similarity between expert-specific features, which are derived by averaging all patch features that selected that expert (via Top-k) within each image, across the MVTec-AD dataset.
The results demonstrate the effectiveness of our specialization strategies, FOFS and ETF Loss. While the baseline exhibits significant redundancy with high similarity scores among experts, our model minimizes this overlap, particularly in the intermediate layers. This confirms that our approach successfully enforces experts to learn distinct functions.

\section{More Visualization Results}
\label{sec:visualization}

\subsection{Multi-layer and Multi-Scale Maps}
\label{sec:multi}
To demonstrate the complementarity of Multi-Level features, we visualize anomaly maps from various layers (6, 12, 18, 24) and scales ($s \in \{1, 3, 5\}$). As shown in \cref{fig:layer_ablation}, we observe that deeper layers excel at highlighting fine-grained details, while larger scales cover broader contexts. Leveraging this synergy, our Multi-Layer \& Multi-Scale Ensemble yields more accurate segmentation results compared to single configuration.

\subsection{Grad-CAM and Patch Selection Map}
\label{sec:gradcam}
\cref{grad-cam(hazelnut)6,grad-cam(hazelnut)12,grad-cam(hazelnut)18,grad-cam(hazelnut)24} visualize the focus region and expert routing for the independent MoE modules integrated at layers 6, 12, 18, and 24, spanning from low-level to high-level features.

\subsection{Final Anomaly Map}
\label{sec:anomalymap}
\cref{fig:m1,fig:m2,fig:m3,fig:m4,fig:m5,fig:m6,fig:m7,fig:m8,fig:m9,fig:m10,fig:m11,fig:m12,fig:m13,fig:m14,fig:m15,fig:m16,fig:m17,fig:m18,fig:m19,fig:m20} illustrate the anomaly map results across multiple categories in industrial and medical datasets.

\section{Limitations and Future Work}
\label{sec:limitations}
Although MoECLIP model achieves high ZSAD performance across 14 benchmark datasets, it possesses the following limitations. 1) The current study primarily focused on enhancing performance by effectively adapting patch features for the ZSAD task. Consequently, the potential for synergistic improvements through advanced text feature adaptation and utilization with the MoECLIP model remains an underexplored area. 2) While MoECLIP provides anomaly maps to visualize anomalous regions, it does not offer explicit explanations as to why the model identifies a specific area as an anomaly. This lack of explainability is a significant drawback, particularly in medical domains. To address these limitations and extend the scope of this research, we plan to investigate replacing the current CLIP-based backbone with a Multimodal LLM-based backbone. This approach aims to construct a model capable of generating text-based explanations for anomalies and explore a more sophisticated language-vision synergy for enhanced generalization performance.

\begin{table*}[t]
\centering
\caption{Class-wise performance results on the MVTec-AD, VisA, BTAD, and DTD-Synthetic datasets.}
\vspace{-2mm}
\footnotesize

\begin{adjustbox}{width=1.2\textwidth, totalheight=0.9\textheight, keepaspectratio}

\begin{tabular}{c c c c c c}
\toprule
\multirow{2}{*}{Dataset} & 
\multirow{2}{*}{Class} &
\multicolumn{2}{c}{Image-Level} &
\multicolumn{2}{c}{Pixel-Level} \\
\cmidrule(lr){3-4} \cmidrule(lr){5-6}
& & AUROC & AP & AUROC & AP \\
\midrule

% ---------------- MVTec (15 classes + 1 average = 16 rows) ----------------
\multirow{16}{*}{MVTec-AD} 
& bottle       & 95.4 & 98.7 & 91.4 & 60.7 \\
& cable        & 76.7 & 86.8 & 82.8 & 16.3 \\
& capsule      & 95.6 & 99.2 & 95.5 & 27.8 \\
& carpet       & 99.9 & 100  & 99.6 & 82.1 \\
& grid         & 99.5 & 99.8 & 98.0 & 33.6 \\
& hazelnut     & 97.4 & 99.7 & 97.9 & 63.2 \\
& leather      & 100  & 100  & 99.4 & 55.8 \\
& metal\_nut   & 93.8 & 98.6 & 77.8 & 30.5 \\
& pill         & 86.9 & 97.4 & 87.9 & 30.0 \\
& screw        & 89.2 & 95.6 & 98.5 & 35.5 \\
& tile         & 99.4 & 99.8 & 98.7 & 70.4 \\
& transistor   & 81.0 & 80.4 & 71.9 & 12.5 \\
& toothbrush   & 96.1 & 98.5 & 95.6 & 34.4 \\
& wood         & 98.6 & 99.6 & 98.0 & 68.2 \\
& zipper       & 98.6 & 99.6 & 96.9 & 58.7 \\
\cmidrule{2-6}
& Average & 93.9 & 96.8 & 92.5 & 45.7 \\

\midrule

% ---------------- VisA (12 classes + 1 average = 13 rows) ----------------
\multirow{13}{*}{VisA}
& candle       & 88.5 & 91.8 & 98.8 & 31.9 \\
& capsules     & 83.4 & 91.8 & 97.3 & 40.3 \\
& cashew       & 84.8 & 92.7 & 97.7 & 21.1 \\
& chewinggum   & 98.6 & 99.8 & 99.5 & 81.5 \\
& fryum        & 90.2 & 96.2 & 94.7 & 30.3 \\
& macaroni1    & 80.6 & 81.3 & 97.3 & 9.8 \\
& macaroni2    & 57.0 & 55.7 & 96.6 & 1.7 \\
& pcb1         & 72.0 & 71.1 & 91.1 & 5.5 \\
& pcb2         & 79.5 & 81.1 & 90.6 & 10.9 \\
& pcb3         & 77.4 & 79.3 & 90.5 & 18.0 \\
& pcb4         & 96.1 & 95.8 & 95.9 & 28.7 \\
& pip\_fryum   & 95.5 & 98.2 & 97.5 & 33.2 \\
\cmidrule{2-6}
& Average & 83.6 & 86.2 & 95.6 & 26.1 \\

\midrule

% ---------------- BTAD (3 classes + 1 average = 4 rows) ----------------
\multirow{4}{*}{BTAD}
& 01 & 98.9 & 99.6 & 96.6 & 53.3 \\
& 02 & 80.6 & 96.8 & 95.8 & 63.1 \\
& 03 & 99.7 & 97.6 & 97.9 & 34.8 \\
\cmidrule{2-6}
& Average & 93.1 & 98.0 & 96.8 & 50.4 \\

\midrule

% ---------------- DTD-Synthetic (12 classes + 1 average = 13 rows) ----------------
\multirow{13}{*}{DTD-Synthetic}
& Blotchy\_099     & 96.5 & 99.1 & 99.3 & 67.6 \\
& Fibrous\_183     & 99.3 & 99.7 & 99.4 & 69.5 \\
& Marbled\_078     & 96.1 & 99.0 & 99.3 & 65.4 \\
& Matted\_069      & 95.3 & 98.8 & 99.1 & 58.3 \\
& Mesh\_114        & 89.7 & 96.0 & 98.0 & 51.8 \\
& Perforated\_037  & 84.5 & 99.2 & 96.9 & 54.5 \\
& Stratified\_154  & 97.2 & 99.3 & 99.6 & 73.9 \\
& Woven\_001  & 99.1 & 99.7 & 99.8 & 69.4 \\
& Woven\_068  & 96.7 & 98.2 & 98.9 & 54.5 \\
& Woven\_104  & 98.1 & 99.3 & 98.5 & 64.1 \\
& Woven\_125  & 99.5 & 99.7 & 99.6 & 69.8 \\
& Woven\_127  & 94.5 & 94.8 & 96.9 & 53.6 \\
\cmidrule{2-6}
& Average & 95.5 & 98.6 & 98.8 & 62.7 \\

\bottomrule
\end{tabular}
\end{adjustbox}
\label{tab:full_results}
\end{table*}

{\small
\begin{figure*}[t]
    \centering
    \vspace{-2.5mm}
    \includegraphics[width=1\linewidth]{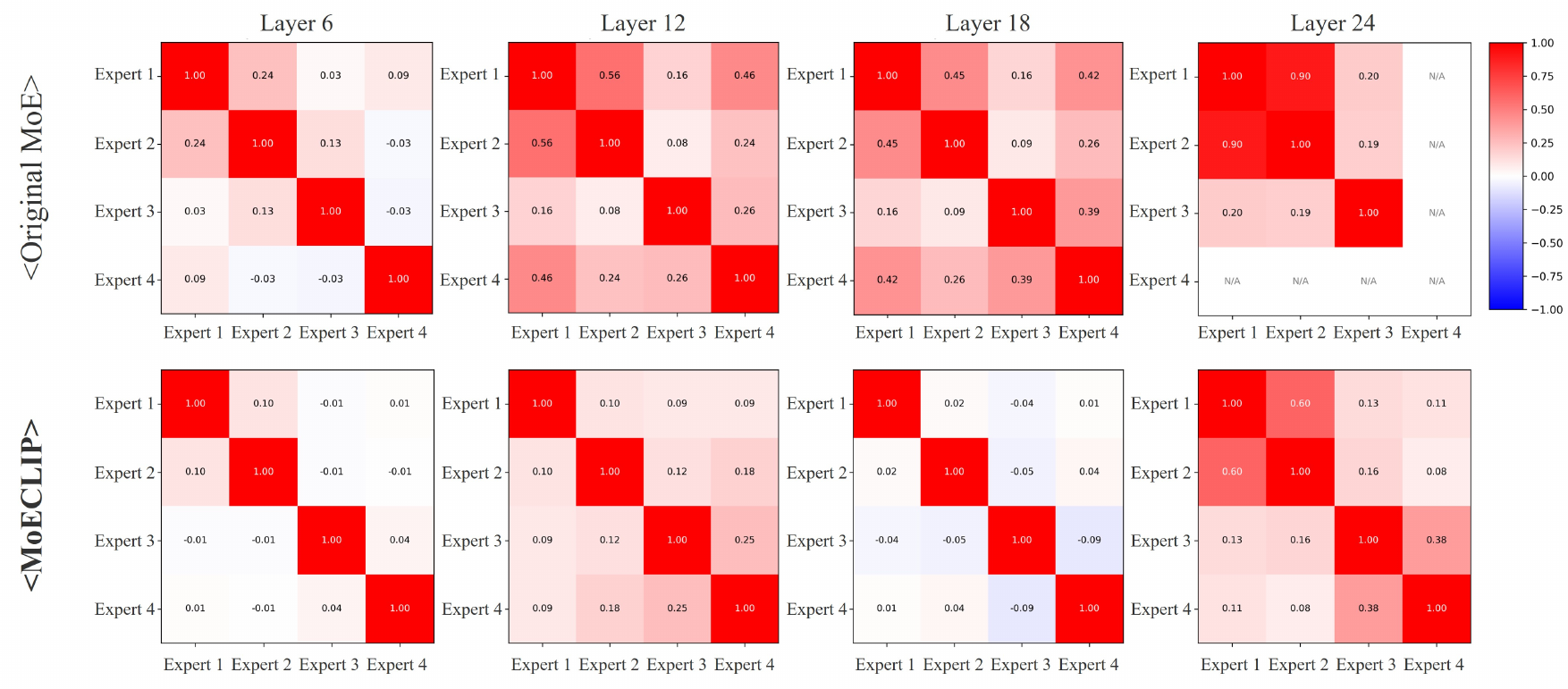}
    \vspace{-4mm}
    \caption{Inter-expert cosine similarity across layers on the MVTec-AD dataset. The top row represents the Original MoE lacking FOFS \&\ ETF Loss, and the bottom row represents our full MoECLIP model. Each column corresponds to a different ViT layer. Values approaching +1 (red) indicate high redundancy, while values approaching 0 (white) or negative values (blue) signify successful differentiation.}
    \label{fig:all_heatmaps}
    \vspace{-2.5mm}
\end{figure*}
}

\begin{figure*}[t!]
    \centering
    \includegraphics[width=1.9\columnwidth]{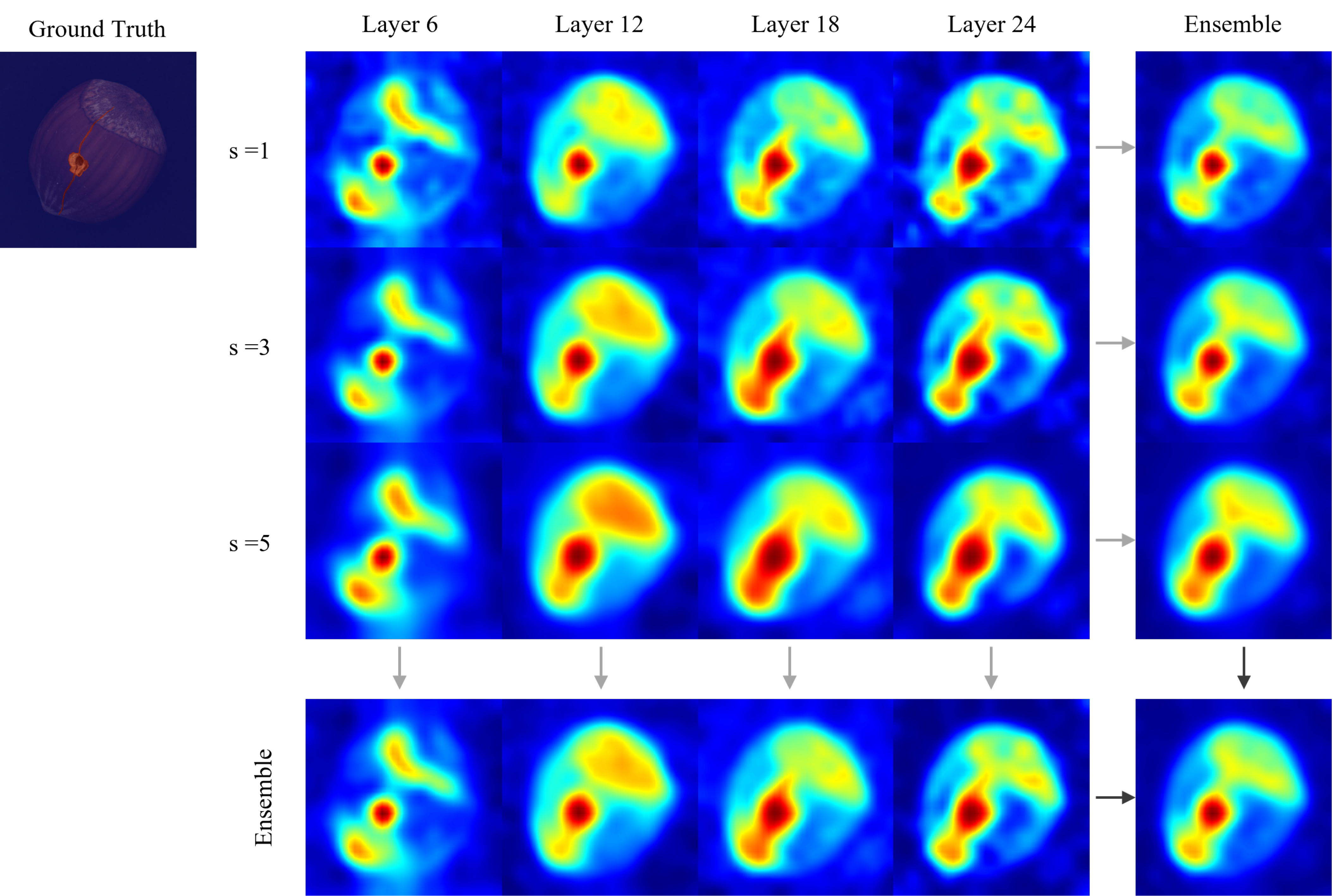}
    \caption{Visualizations from different scales ($s \in \{1, 3, 5\}$) and layers (6, 12, 18, 24) demonstrate complementary characteristics across depth and spatial context. The final ensemble (bottom-right) presents a unified anomaly map combining all scales and layers.}
    \label{fig:layer_ablation}
\end{figure*}

\begin{figure*}
    \centering
    \includegraphics[width=1.9\columnwidth]{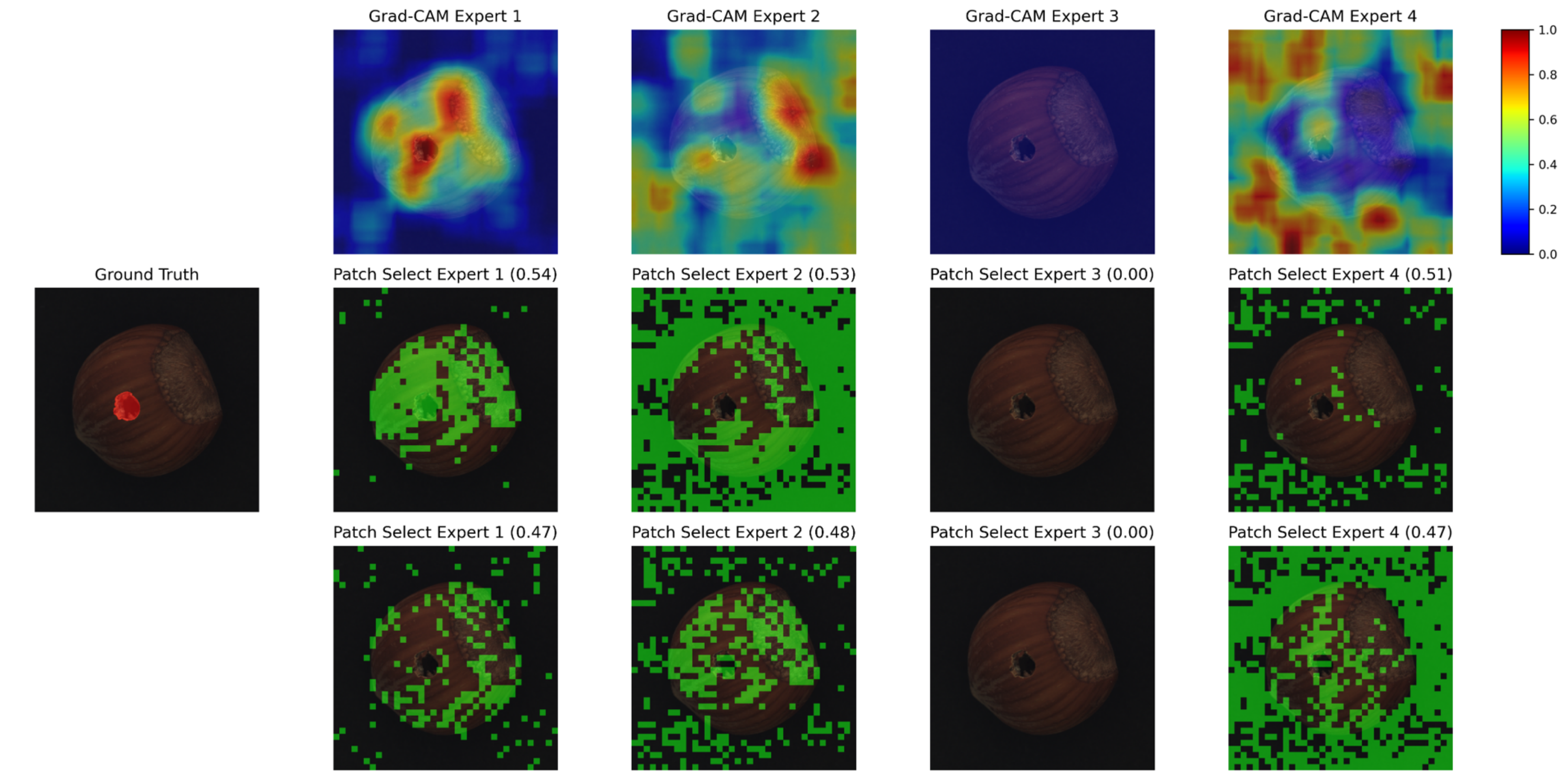}
    \caption{\textbf{Visualization of Grad-CAM and patch selection maps for each expert at layer 6 on the hazelnut class of the MVTec-AD dataset.} The Ground Truth image is shown on the far left. The first row (Grad-CAM) highlights each expert’s focus region. The second and third rows (Patch Selection) illustrate the patches where the corresponding expert was selected as the router’s Top-1 and Top-2 choices, respectively (shown in green). The value in each subplot title represents the expert’s average renormalized routing weight computed from its Top-1 and Top-2 assigned patches under their respective routing settings.}
    \label{grad-cam(hazelnut)6}
\end{figure*}

\begin{figure*}
    \centering
    \includegraphics[width=1.9\columnwidth]{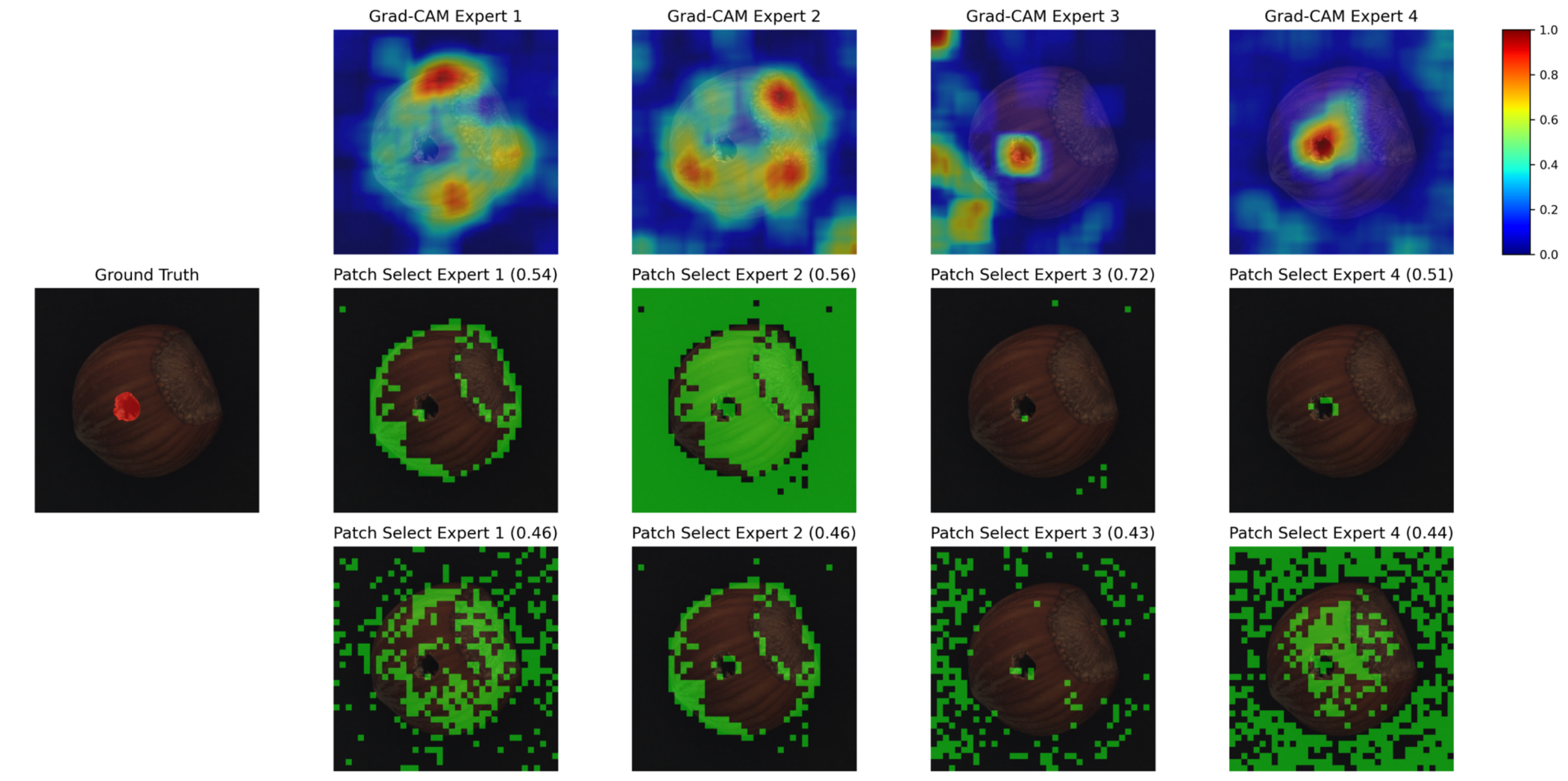}
    \caption{\textbf{Visualization of Grad-CAM and patch selection maps for each expert at layer 12 on the hazelnut class of the MVTec-AD dataset.} The Ground Truth image is shown on the far left. The first row (Grad-CAM) highlights each expert’s focus region. The second and third rows (Patch Selection) illustrate the patches where the corresponding expert was selected as the router’s Top-1 and Top-2 choices, respectively (shown in green). The value in each subplot title represents the expert’s average renormalized routing weight computed from its Top-1 and Top-2 assigned patches under their respective routing settings.}
    \label{grad-cam(hazelnut)12}
\end{figure*}

\begin{figure*}
    \centering
    \includegraphics[width=1.9\columnwidth]{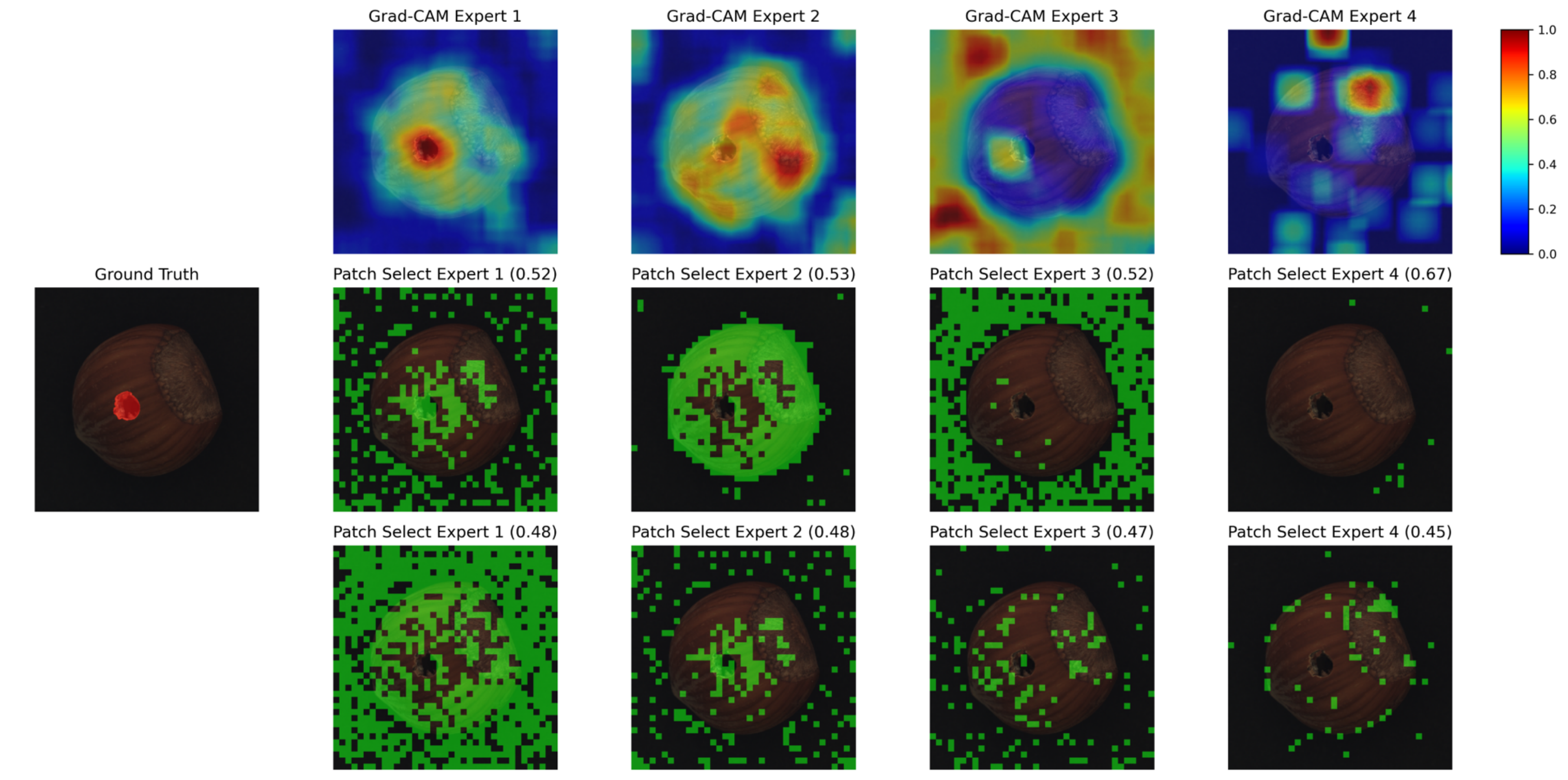}
    \caption{\textbf{Visualization of Grad-CAM and patch selection maps for each expert at layer 18 on the hazelnut class of the MVTec-AD dataset.} The Ground Truth image is shown on the far left. The first row (Grad-CAM) highlights each expert’s focus region. The second and third rows (Patch Selection) illustrate the patches where the corresponding expert was selected as the router’s Top-1 and Top-2 choices, respectively (shown in green). The value in each subplot title represents the expert’s average renormalized routing weight computed from its Top-1 and Top-2 assigned patches under their respective routing settings.}
    \label{grad-cam(hazelnut)18}
\end{figure*}

\begin{figure*}
    \centering
    \includegraphics[width=1.9\columnwidth]{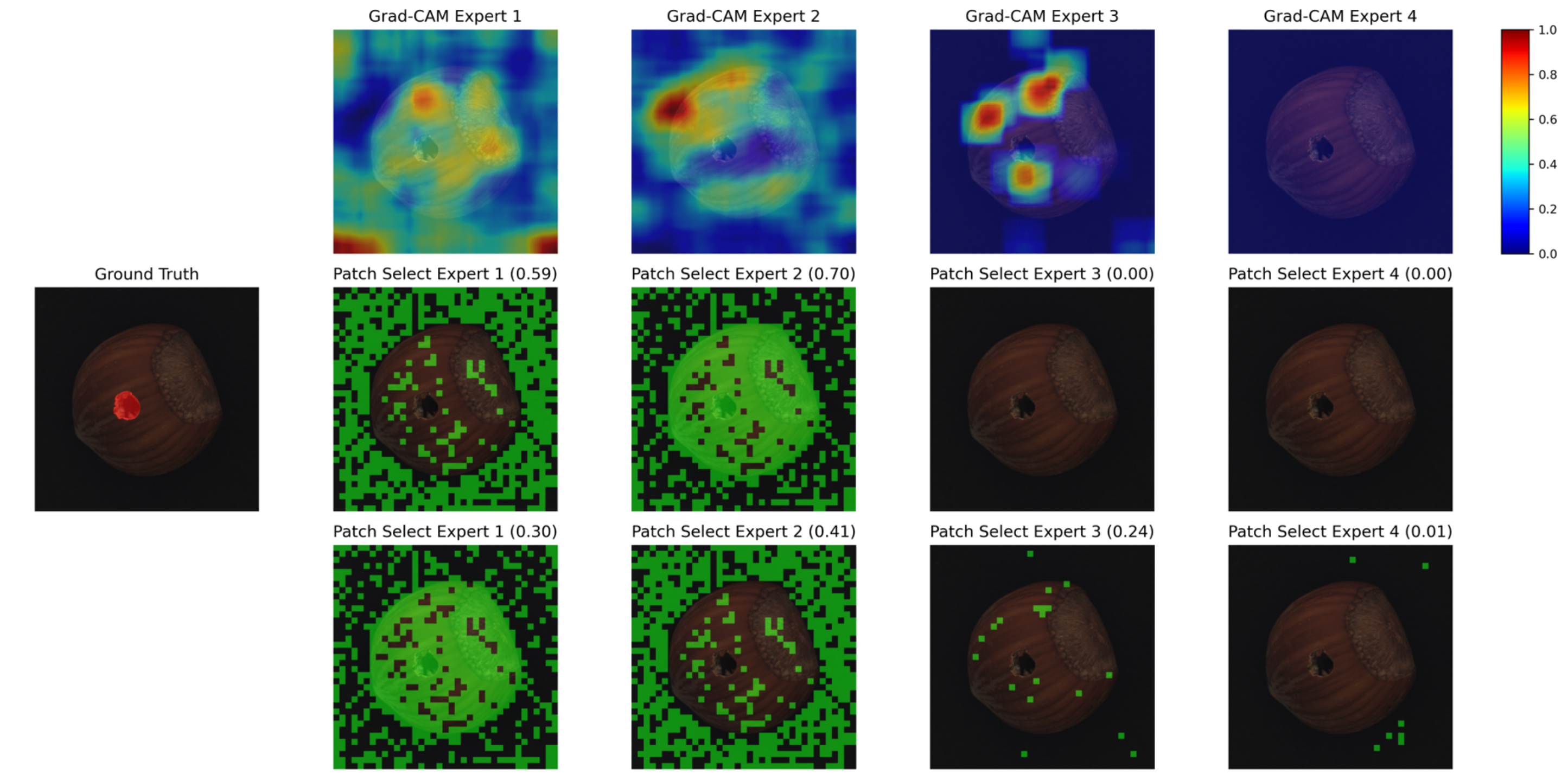}
    \caption{\textbf{Visualization of Grad-CAM and patch selection maps for each expert at layer 24 on the hazelnut class of the MVTec-AD dataset.} The Ground Truth image is shown on the far left. The first row (Grad-CAM) highlights each expert’s focus region. The second and third rows (Patch Selection) illustrate the patches where the corresponding expert was selected as the router’s Top-1 and Top-2 choices, respectively (shown in green). The value in each subplot title represents the expert’s average renormalized routing weight computed from its Top-1 and Top-2 assigned patches under their respective routing settings.}
    \label{grad-cam(hazelnut)24}
\end{figure*}

\begin{figure*}
    \centering
    \includegraphics[width=1.9\columnwidth]{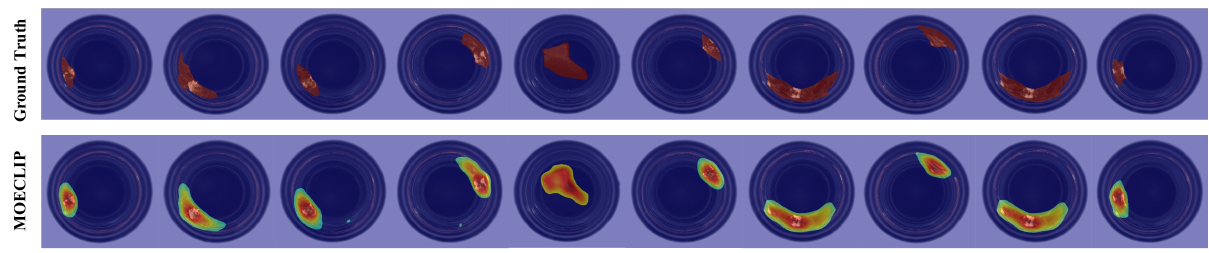}
    \caption{\textbf{Anomaly Map results for the bottle in the MVTec-AD.}
The first row contains the original images with red areas showing the ground truth. The second row shows the anomaly map results generated by MoECLIP.}
    \label{fig:m1}
\end{figure*}

\begin{figure*}
    \centering
    \includegraphics[width=1.9\columnwidth]{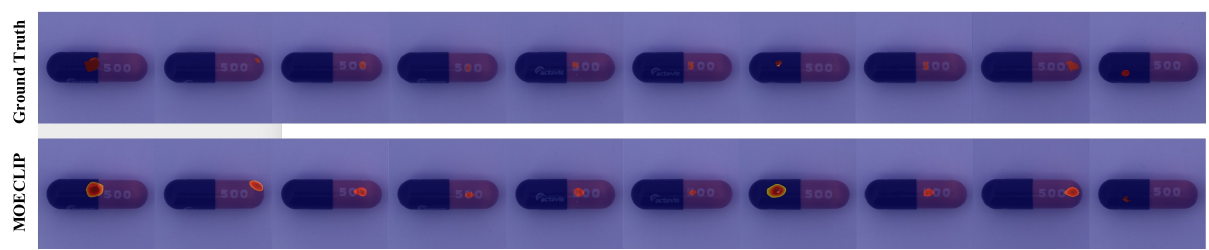}
    \caption{\textbf{Anomaly Map results for the capsule in the MVTec-AD.}
The first row contains the original images with red areas showing the ground truth. The second row shows the anomaly map results generated by MoECLIP.}
    \label{fig:m2}
\end{figure*}

\begin{figure*}
    \centering
    \includegraphics[width=1.9\columnwidth]{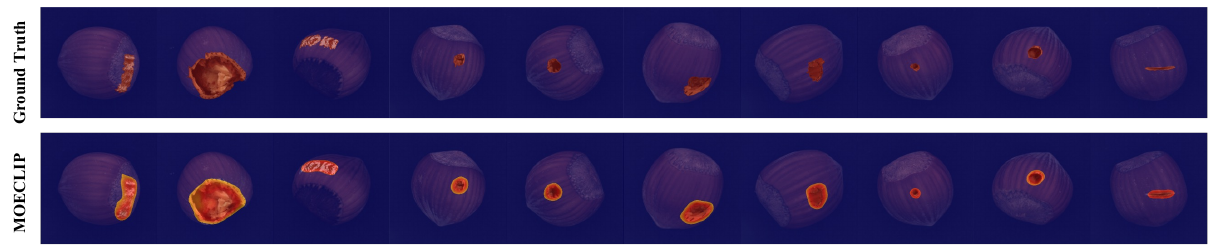}
    \caption{\textbf{Anomaly Map results for the hazelnut in the MVTec-AD.}
The first row contains the original images with red areas showing the ground truth. The second row shows the anomaly map results generated by MoECLIP.}
    \label{fig:m3}
\end{figure*}

\begin{figure*}
    \centering
    \includegraphics[width=1.9\columnwidth]{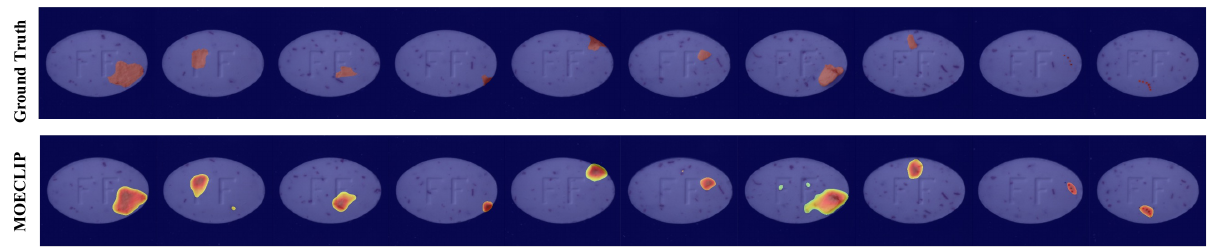}
    \caption{\textbf{Anomaly Map results for the pill in the MVTec-AD.}
The first row contains the original images with red areas showing the ground truth. The second row shows the anomaly map results generated by MoECLIP.}
    \label{fig:m4}
\end{figure*}

\begin{figure*}
    \centering
    \includegraphics[width=1.9\columnwidth]{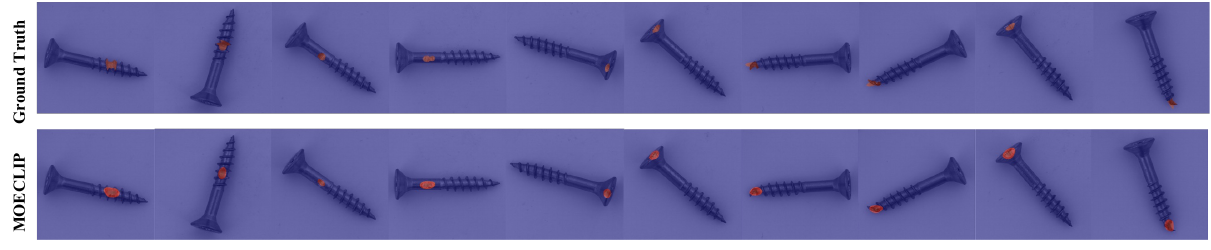}
    \caption{\textbf{Anomaly Map results for the screw in the MVTec-AD.}
The first row contains the original images with red areas showing the ground truth. The second row shows the anomaly map results generated by MoECLIP.}
    \label{fig:m5}
\end{figure*}

\begin{figure*}
    \centering
    \includegraphics[width=1.9\columnwidth]{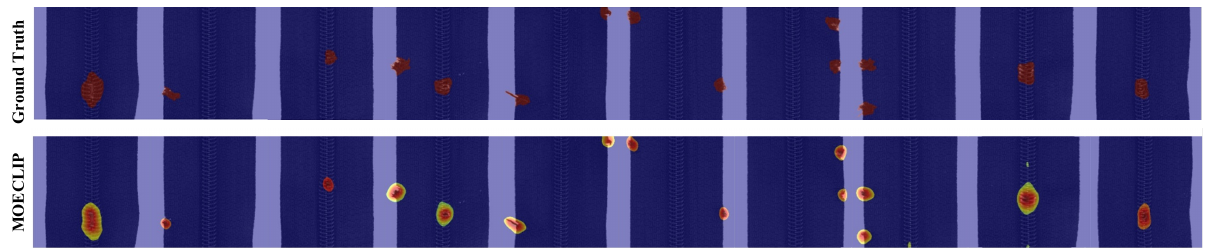}
    \caption{\textbf{Anomaly Map results for the zipper in the MVTec-AD.}
The first row contains the original images with red areas showing the ground truth. The second row shows the anomaly map results generated by MoECLIP.}
    \label{fig:m6}
\end{figure*}

\begin{figure*}
    \centering
    \includegraphics[width=1.9\columnwidth]{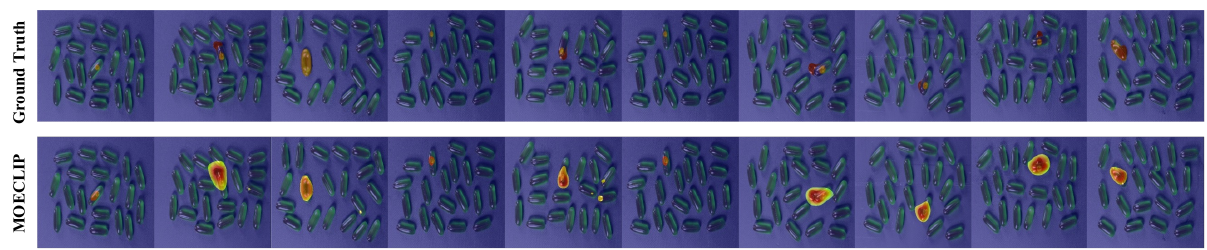}
    \caption{\textbf{Anomaly Map results for the capsules in the VisA.}
The first row contains the original images with red areas showing the ground truth. The second row shows the anomaly map results generated by MoECLIP.}
    \label{fig:m7}
\end{figure*}

\begin{figure*}
    \centering
    \includegraphics[width=1.9\columnwidth]{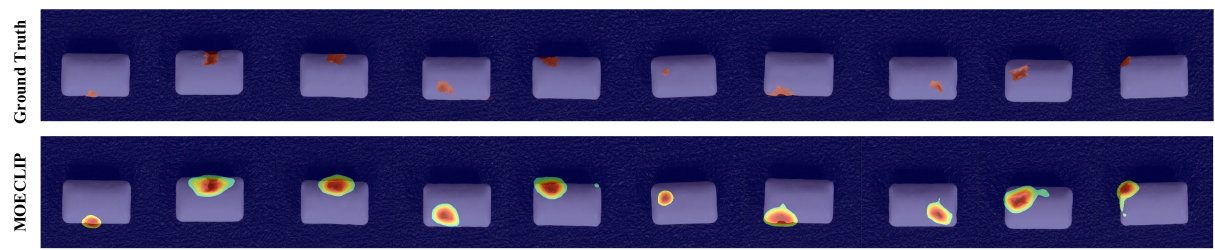}
    \caption{\textbf{Anomaly Map results for the chewinggum in the VisA.}
The first row contains the original images with red areas showing the ground truth. The second row shows the anomaly map results generated by MoECLIP.}
    \label{fig:m8}
\end{figure*}

\begin{figure*}
    \centering
    \includegraphics[width=1.9\columnwidth]{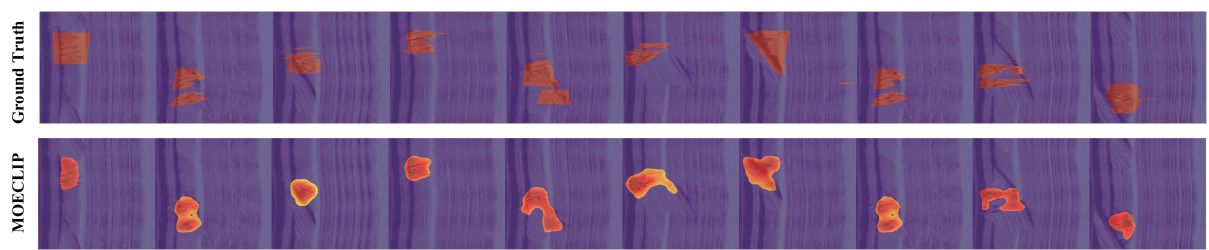}
    \caption{\textbf{Anomaly Map results for the 02 in the BTAD.}
The first row contains the original images with red areas showing the ground truth. The second row shows the anomaly map results generated by MoECLIP.}
    \label{fig:m9}
\end{figure*}
\begin{figure*}
    \centering
    \includegraphics[width=1.9\columnwidth]{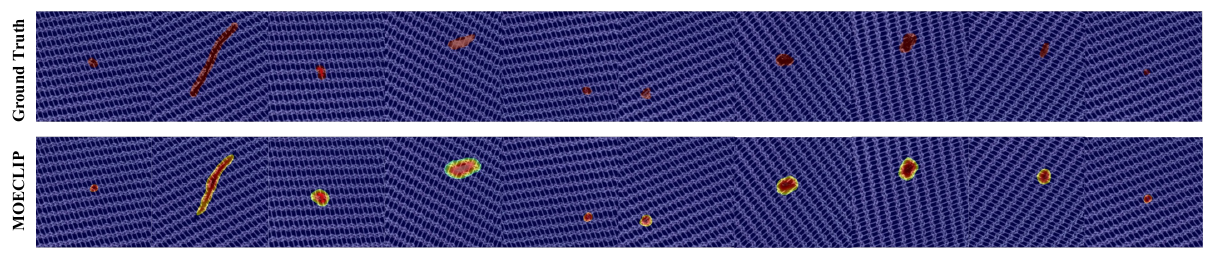}
    \caption{\textbf{Anomaly Map results for the Mesh\_114 in the DTD-Synthetic.}
The first row contains the original images with red areas showing the ground truth. The second row shows the anomaly map results generated by MoECLIP.}
    \label{fig:m10}
\end{figure*}

\begin{figure*}
    \centering
    \includegraphics[width=1.9\columnwidth]{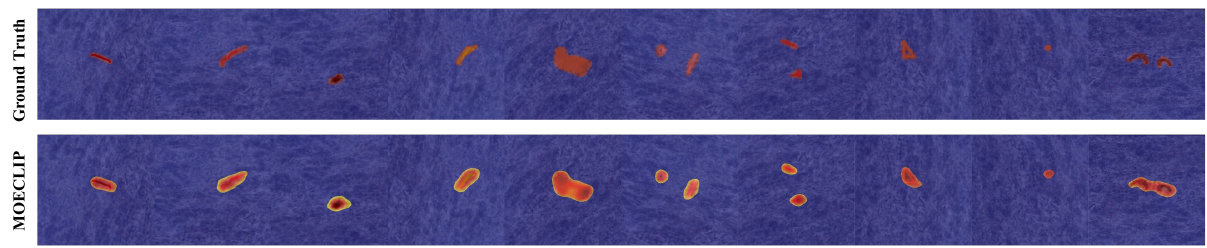}
    \caption{\textbf{Anomaly Map results for the Marbled\_079 in the DTD-Synthetic.}
The first row contains the original images with red areas showing the ground truth. The second row shows the anomaly map results generated by MoECLIP.}
    \label{fig:m11}
\end{figure*}

\begin{figure*}
    \centering
    \includegraphics[width=1.9\columnwidth]{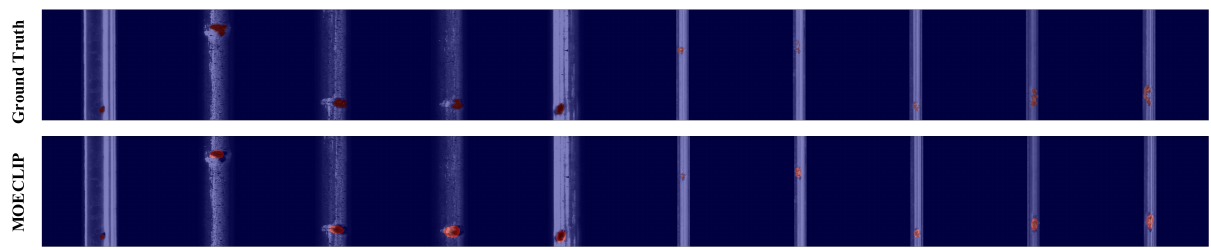}
    \caption{\textbf{Anomaly Map results on the RSDD.}
The first row contains the original images with red areas showing the ground truth. The second row shows the anomaly map results generated by MoECLIP.}
    \label{fig:m12}
\end{figure*}

\begin{figure*}
    \centering
    \includegraphics[width=1.9\columnwidth]{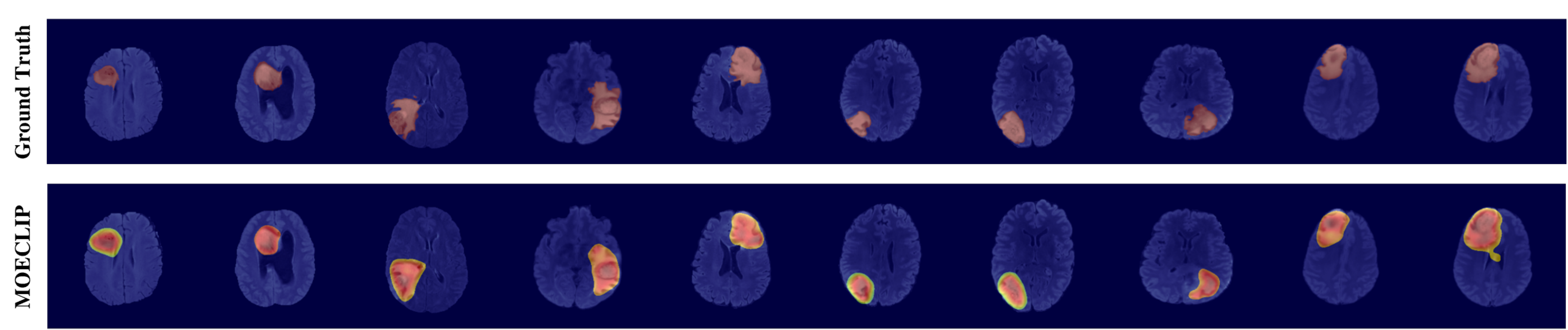}
    \caption{\textbf{Anomaly Map results on the Brain MRI.}
The first row contains the original images with red areas showing the ground truth. The second row shows the anomaly map results generated by MoECLIP.}
    \label{fig:m13}
\end{figure*}

\begin{figure*}
    \centering
    \includegraphics[width=1.9\columnwidth]{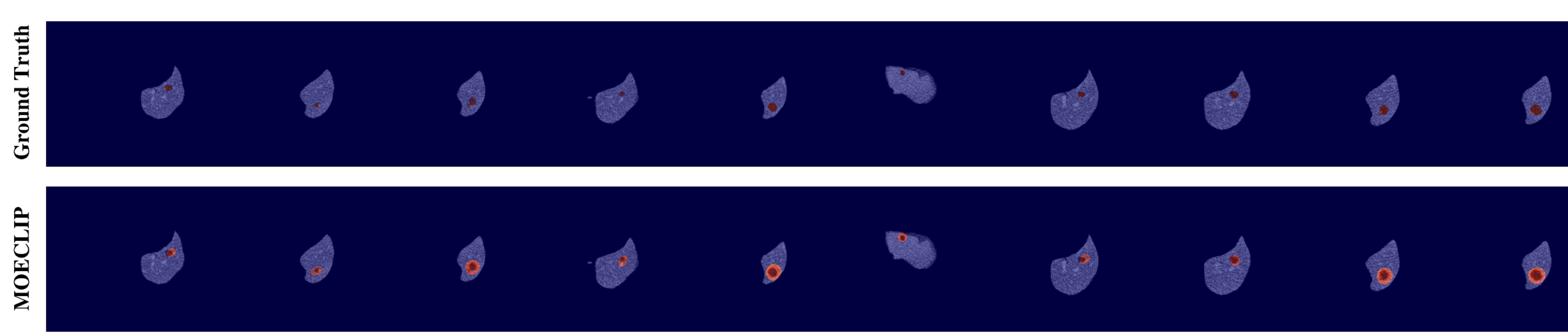}
    \caption{\textbf{Anomaly Map results on the Liver CT.}
The first row contains the original images with red areas showing the ground truth. The second row shows the anomaly map results generated by MoECLIP.}
    \label{fig:m14}
\end{figure*}

\begin{figure*}
    \centering
    \includegraphics[width=1.9\columnwidth]{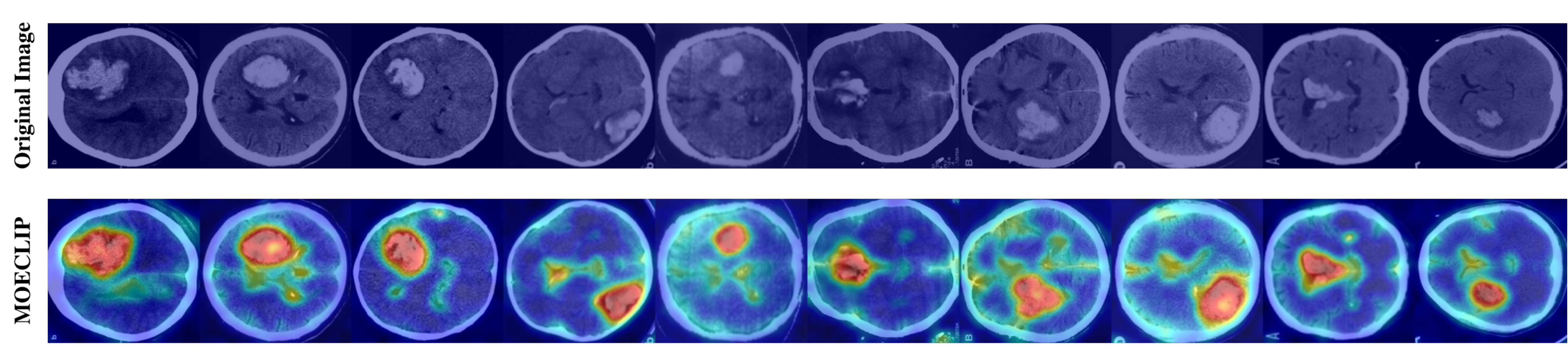}
    \caption{\textbf{Anomaly Map results on the Head CT.}
The absence of pixel-level annotations restricts the use of this dataset to anomaly classification. The first row contains the original images. The second row shows the anomaly map results generated by MoECLIP.}
    \label{fig:m15}
\end{figure*}

\begin{figure*}
    \centering
    \includegraphics[width=1.9\columnwidth]{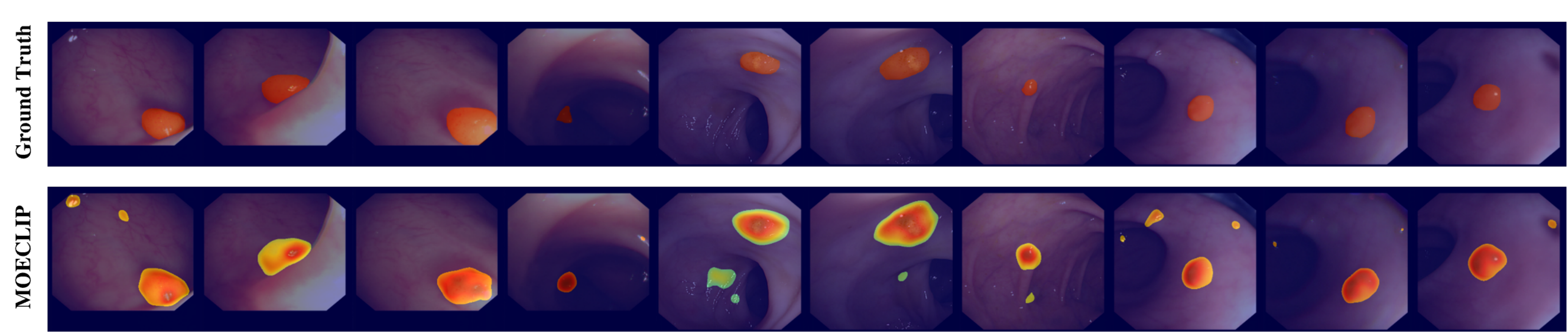}
    \caption{\textbf{Anomaly Map results on the CVC-ColonDB.}
The first row contains the original images with red areas showing the ground truth. The second row shows the anomaly map results generated by MoECLIP.}
    \label{fig:m16}
\end{figure*}

\begin{figure*}
    \centering
    \includegraphics[width=1.9\columnwidth]{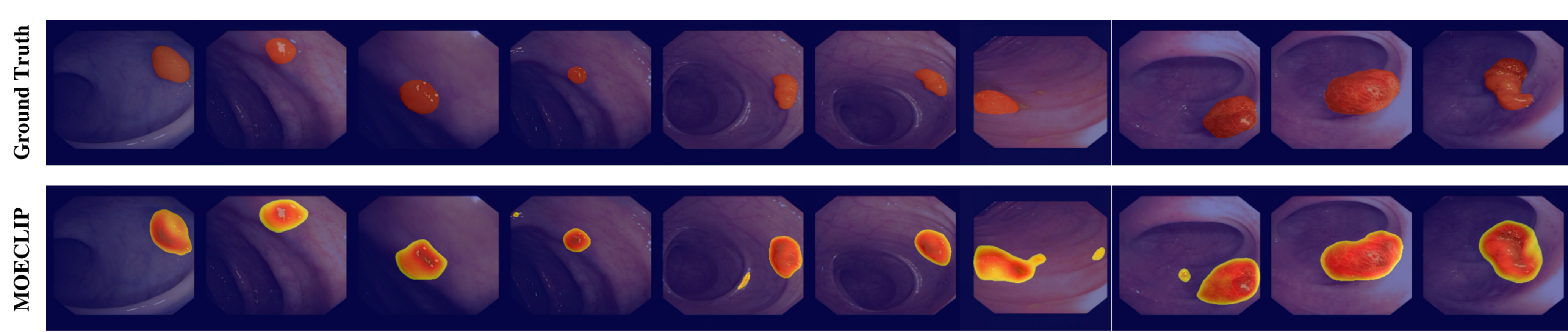}
    \caption{\textbf{Anomaly Map results on the CVC-ClinicDB.}
The first row contains the original images with red areas showing the ground truth. The second row shows the anomaly map results generated by MoECLIP.}
    \label{fig:m17}
\end{figure*}

\begin{figure*}
    \centering
    \includegraphics[width=1.9\columnwidth]{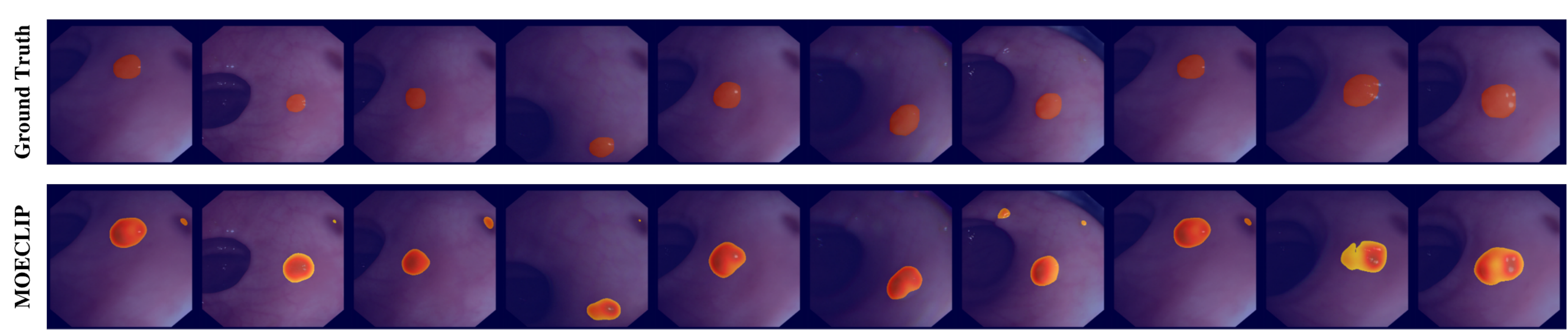}
    \caption{\textbf{Anomaly Map results on the CVC-300.}
The first row contains the original images with red areas showing the ground truth. The second row shows the anomaly map results generated by MoECLIP.}
    \label{fig:m18}
\end{figure*}

\begin{figure*}
    \centering
    \includegraphics[width=1.9\columnwidth]{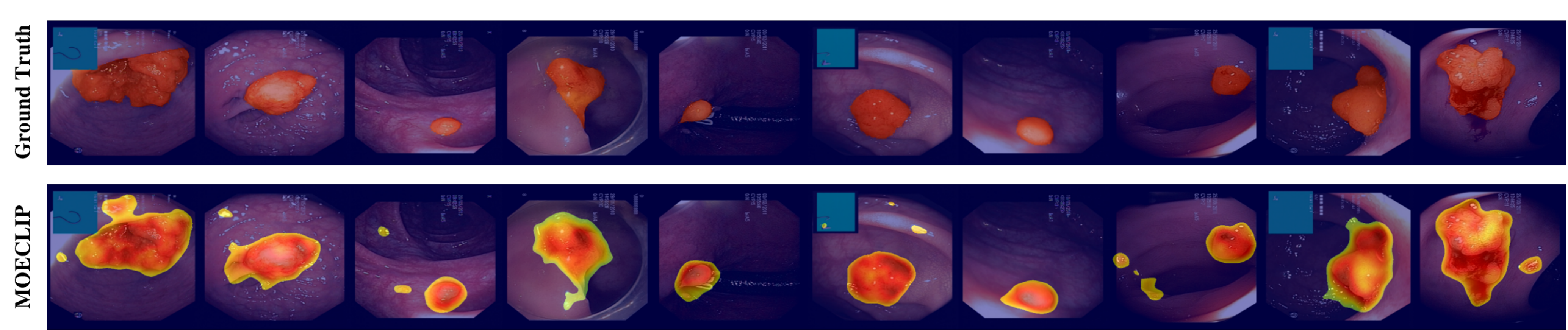}
    \caption{\textbf{Anomaly Map results on the Endo.}
The first row contains the original images with red areas showing the ground truth. The second row shows the anomaly map results generated by MoECLIP.}
    \label{fig:m19}
\end{figure*}

\begin{figure*}
    \centering
    \includegraphics[width=1.9\columnwidth]{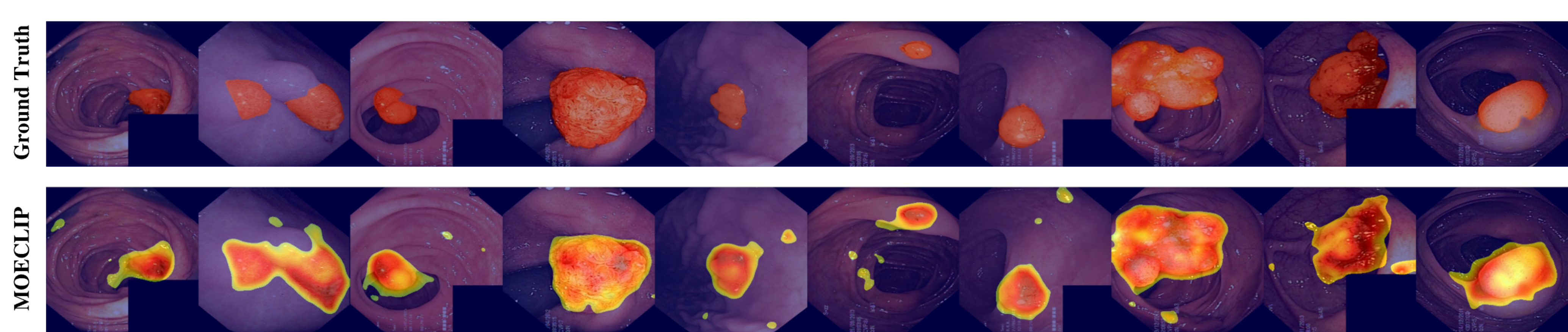}
    \caption{\textbf{Anomaly Map results on the Kvasir.}
The first row contains the original images with red areas showing the ground truth. The second row shows the anomaly map results generated by MoECLIP.}
    \label{fig:m20}
\end{figure*}


\end{document}